\newif\ifversionA
\versionAfalse

\documentclass{article}

\PassOptionsToPackage{numbers, compress}{natbib}

\usepackage[main,final]{neurips_2026}

\usepackage[utf8]{inputenc} 
\usepackage[T1]{fontenc}    
\usepackage{hyperref}       
\usepackage{url}            
\usepackage{booktabs}       
\usepackage{amsfonts}       
\usepackage{nicefrac}       
\usepackage{microtype}      
\usepackage{amssymb}
\usepackage{graphicx}

\usepackage{multirow}
\usepackage{array}
\usepackage{xcolor,colortbl}
\usepackage{arydshln}
\usepackage{bm}
\usepackage{wrapfig}
\usepackage{amsmath}
\usepackage{tabularx}
\usepackage[table]{xcolor}
\usepackage[normalem]{ulem}
\definecolor{softlavender}{RGB}{235,233,246}
\newcommand{\splitsep}{\color{black!60}\vrule width 0.8pt}
\newcommand{\datasetsep}{\color{black}\vrule width 1.0pt}
\newcolumntype{C}[1]{>{\centering\arraybackslash}p{#1}}

\newcommand{\best}[1]{\textcolor{red}{\textbf{#1}}}
\newcommand{\second}[1]{\textcolor{blue}{\underline{#1}}}

\hypersetup{
    colorlinks=true,
    citecolor={blue!50!black},
    linkcolor={red!50!black},
    urlcolor=black
}

\title{PoseBridge: Bridging the Skeletonization Gap for Zero-Shot Skeleton-Based Action Recognition}

\author{%
  Sanghyeon Lee \quad Jinwoo Kim \quad Jong Taek Lee \\
  School of Computer Science and Engineering\\
  Kyungpook National University, Daegu, South Korea \\
  \texttt{\{hyeon1263,kk234556,jongtaeklee\}@knu.ac.kr} \\
}

\begin{document}

\maketitle

\begin{abstract}  
  Zero-shot skeleton-based action recognition (ZSSAR) is typically treated as a skeleton-text alignment problem: encode joint-coordinate sequences, align them with language, and classify unseen actions. We argue that this alignment is often too late. Skeletons are not complete action observations, but compressed outputs of human pose estimation (HPE); by the time alignment begins, human-object interactions and pose-relative visual cues may no longer be explicit. We call this upstream semantic loss. To address it, we propose \textbf{PoseBridge}, an HPE-aware ZSSAR framework that bridges intermediate HPE representations to skeleton-text alignment. Rather than adding an RGB action branch or object detector, PoseBridge extracts pose-anchored semantic cues from the same HPE process that produces skeletons, then transfers them through skeleton-conditioned bridging and semantic prototype adaptation. Across NTU-RGB+D 60/120, PKU-MMD, and Kinetics-200/400, PoseBridge improves ZSSAR performance under the evaluated protocols. On the Kinetics-200/400 PURLS benchmark, which contains in-the-wild videos with diverse scenes and action contexts, PoseBridge achieves substantial gains over prior ZSSAR methods, outperforming the strongest compared baseline by 13.3--17.4 points across all eight splits. Code is available at \url{https://github.com/hyeon0819/PoseBridge}.
\end{abstract}

\section{Introduction}
\label{sec:introduction}

Skeleton-based action recognition represents actions as compact body-joint trajectories, providing motion-centric cues less tied to appearance variation, background clutter, and illumination changes than raw RGB observations~\cite{yan2018spatial,shi2019two,duan2022revisiting}. 
These properties are particularly attractive for zero-shot skeleton-based action recognition (ZSSAR), where models recognize unseen actions by transferring knowledge through shared motion patterns and textual action descriptions~\cite{jasani2019skeleton,9506179,Zhu_2024_CVPR,Do_2025_ICCV, kuang2025zero}. 
Most ZSSAR methods~\cite{jasani2019skeleton,9506179,zhou2023zero,Zhu_2024_CVPR,Do_2025_ICCV,Zhu_2025_CVPR} follow a skeletonize-then-align (S2A) paradigm: raw visual observations are first reduced to joint coordinates by a human pose estimation (HPE) model, and the resulting skeleton sequences are then aligned with text features in a shared semantic space.

\begin{figure*}[t]
\begin{center}
\includegraphics[width=1.0\textwidth]{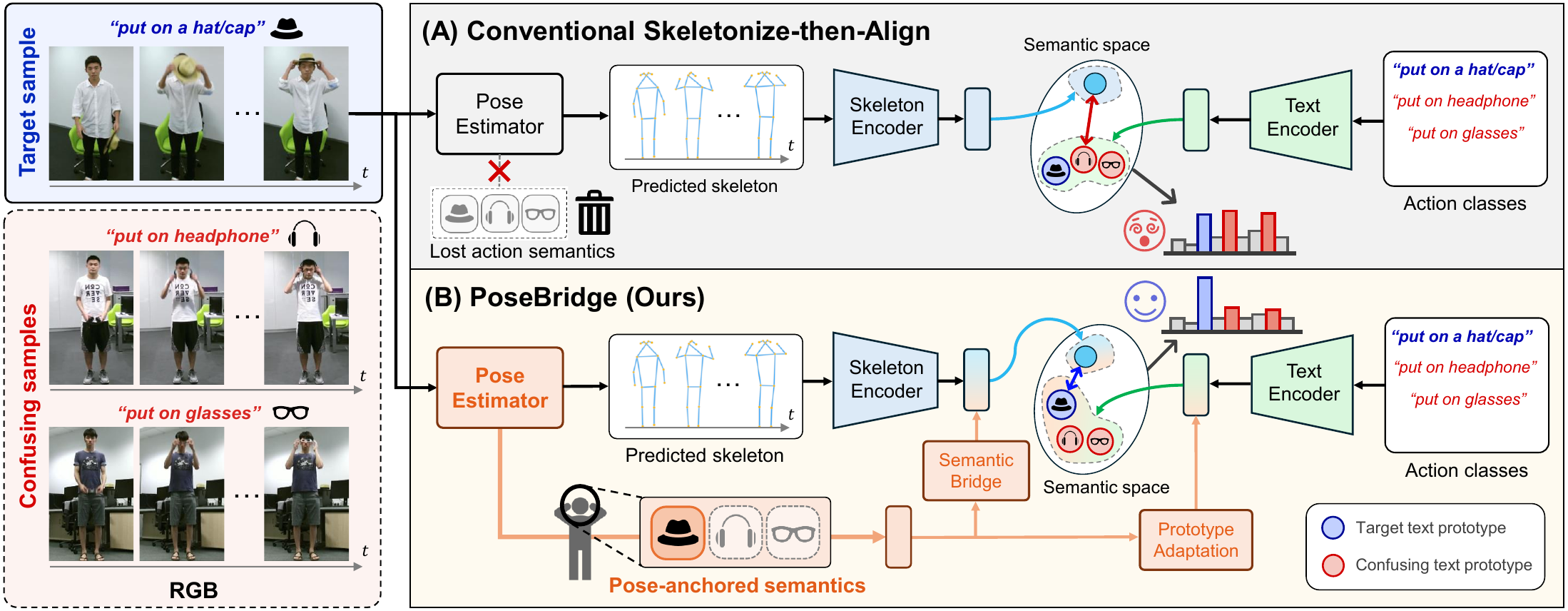}
\end{center}
\caption{Motivation of PoseBridge.
(A) Conventional S2A aligns text with joint-coordinate skeletons after skeletonization, where object, interaction, and pose-relative cues are not explicit. 
(B) PoseBridge routes HPE-side cues to the alignment space, disambiguating object-sensitive actions with similar skeleton trajectories.
}
\label{fig:figure1}
\end{figure*}

However, this paradigm relies on a rarely examined assumption: skeletons preserve sufficient action semantics for language alignment. In practice, a skeleton is a compressed output of an upstream pose estimator rather than a complete action observation. When raw visual evidence is converted into joint coordinates, cues beyond body motion---such as object presence, human-object interaction, and pose-relative object relations---are no longer explicit. Object-sensitive benchmark actions expose this limitation: as shown in Fig.~\ref{fig:figure1}(A), ``put on a hat/cap'', ``put on headphone'', and ``put on glasses'' can share similar hand-to-head trajectories, but differ in the manipulated object and its configuration with the actor's hands and head. As a result, S2A may place similar skeleton features near multiple text prototypes, producing ambiguous zero-shot predictions. We call this mismatch between the semantics required by action text and the evidence retained in skeletons the skeletonization gap, and identify its source as upstream semantic loss before skeleton-text alignment begins.

A direct way to mitigate this gap is to augment skeletons with additional context. Language- or prompt-based context can enrich skeleton-text alignment with object and scene descriptions~\cite{Chen_2025_CVPR,wang2026skeletoncontext}, but such context is derived from language or skeleton-side inference rather than recovered from the visual evidence available inside the HPE stream. Another option is to introduce an RGB or object-centric visual pathway~\cite{liu2025beyond,sinha2025ski}. However, this changes the modality protocol of ZSSAR and typically requires an additional visual representation beyond the HPE pass used to obtain skeletons, while also exposing the model to appearance or scene shortcuts. Reusing HPE features is closer to the skeleton extraction process~\cite{wen2025enhancing}, but naive feature fusion treats intermediate maps as generic visual features and does not ensure that the retained cues are action-relevant or organized around the human body. The key challenge is therefore not merely to reuse HPE features or append generic context after skeletonization, but to explicitly preserve action-relevant visual evidence within HPE and learn how it should be transferred to skeleton-based recognition. This shifts the focus from improving skeleton-text alignment after information has been discarded to addressing the upstream information bottleneck itself.

To address this challenge, we propose \textbf{PoseBridge}, a pose-anchored semantic bridge framework for ZSSAR. As shown in Fig.~\ref{fig:figure1}(B), PoseBridge intervenes before skeleton-text alignment by deriving bridge cues from intermediate HPE representations produced during skeleton extraction. Instead of treating HPE as a preprocessing step that only outputs joint coordinates, PoseBridge reinterprets its intermediate representations as pose-anchored semantic cues grounded in the body structure that defines the skeleton. These cues provide object- and pose-relative evidence that joint-coordinate skeletons do not explicitly retain. Importantly, PoseBridge uses cues from the same HPE process that produces skeletons, without an independent RGB action-recognition branch or external object detector.

PoseBridge consists of two coupled stages. First, it learns pose-anchored semantic cues within HPE. Multi-level HPE features are hierarchically refined so that fine-grained visual evidence from shallow layers is injected into deeper pose-structured representations, and body-aware pooling aggregates the refined responses around predicted joints. Auxiliary semantic alignment further encourages the pooled cues to preserve action-relevant evidence rather than generic appearance. Second, PoseBridge transfers these cues to ZSSAR. A skeleton-conditioned semantic bridge lets motion representations query temporal pose-anchored cues, while semantic prototype adaptation calibrates language prototypes toward the pose-semantic space estimated from seen classes. Together, these two stages establish a unified preservation-and-transfer framework, enabling action-relevant visual evidence to be retained during skeleton extraction and effectively exploited for unseen-action recognition.

Experiments evaluate PoseBridge on NTU-RGB+D 60/120, PKU-MMD, and the broader Kinetics-200/400 PURLS benchmark, covering zero-shot learning (ZSL), generalized ZSL harmonic mean (GZSL-H), and random-split protocols. Beyond the commonly used NTU/PKU-style benchmarks, Kinetics further evaluates ZSSAR under greater class diversity and in-the-wild video conditions, where PoseBridge achieves particularly large improvements over previous methods evaluated under the PURLS protocol.

Our contributions are summarized as follows:
\begin{itemize}
    \item We identify the skeletonization gap as a fundamental limitation of conventional skeletonize-then-align ZSSAR, revealing that action-relevant object, interaction, and pose-relative cues can become non-explicit before skeleton-text alignment begins. This motivates addressing semantic loss at its source rather than only refining downstream alignment.
    
    \item We propose PoseBridge, a pose-anchored semantic bridge framework that addresses this bottleneck through two coupled stages: semantic preservation within HPE via hierarchical refinement and body-aware pooling, and semantic transfer to ZSSAR via skeleton-conditioned bridging and prototype adaptation.

    \item PoseBridge achieves state-of-the-art ZSL accuracy across all standard NTU-RGB+D 60/120 splits, improves random-split GZSL-H by 11.9/12.9 points on NTU-RGB+D 120/PKU-MMD, and surpasses prior ZSSAR methods by 13.3--17.4 points across eight Kinetics-200/400 PURLS splits.
\end{itemize}

\section{Related Work}
\label{sec:related_work}
\ifversionA
\else
\textbf{Skeleton-text alignment for ZSSAR}.\; 
Zero-shot skeleton-based action recognition (ZSSAR) recognizes unseen actions by aligning skeleton representations with textual class semantics~\cite{chen2024fine, Schonfeld_2019_CVPR_Workshops, Wu_2025_ICCV, Zhu_2025_CVPR}. Early methods learn skeleton-text compatibility metrics, e.g., Relation Network~\cite{jasani2019skeleton}, while later methods improve the semantic structure of this alignment. SynSE~\cite{9506179} decomposes action labels into verb and noun components, and SMIE~\cite{zhou2023zero} uses mutual information estimation for cross-modal matching. Recent methods further refine alignment granularity and robustness: PURLS~\cite{Zhu_2024_CVPR} aligns part-level skeleton features with textual descriptions, SA-DVAE~\cite{li2024sa} separates action-relevant and irrelevant skeleton components, TDSM~\cite{Do_2025_ICCV} denoises skeleton features toward descriptive latent representations, and Flora~\cite{chen2025learning} improves semantic modeling with distribution-aware classification. These methods differ in alignment objectives and semantic representations, but largely operate after skeletonization: joint-coordinate sequences are treated as the primary observation, and semantic alignment is performed only after visual evidence has been compressed into skeletons.
\fi

\ifversionA
\else
\textbf{Contextual and pose-estimation cues for skeleton recognition}.\;
Joint-coordinate skeletons encode body motion but do not explicitly represent objects, interactions, or scene context. Language-side methods enrich skeleton recognition with external or generated descriptions: Neuron~\cite{Chen_2025_CVPR} uses LLM-generated contextual side information, and SkeletonContext~\cite{wang2026skeletoncontext} injects prompt-based object and environment cues. These methods improve class-level semantics, but their context is text-derived and cannot directly recover sample-specific visual evidence made non-explicit during skeleton extraction. Other methods introduce visual information to complement skeletons, such as RGB-skeleton fusion in BSZSL~\cite{liu2025beyond} and vision-language alignment in SKI-Models~\cite{sinha2025ski}; however, such approaches change the modality protocol by adding an additional visual pathway. A related line uses pose-estimation-side representations rather than joint coordinates alone, including pose estimation maps~\cite{liu2018recognizing}, heatmap volumes~\cite{duan2022revisiting}, and intermediate HPE feature maps in PFMESR~\cite{wen2025enhancing}. 
These studies highlight the value of HPE representations beyond joint coordinates. In particular, PFMESR reuses intermediate features from a pretrained HPE network for supervised action recognition. However, such reuse does not explicitly preserve visual-language cues for unseen actions or transfer them to zero-shot skeleton-text alignment. In contrast, PoseBridge learns pose-anchored semantic cues within HPE and transfers them through skeleton-conditioned attention and semantic prototype adaptation. Thus, our contribution lies in coupling upstream semantic preservation with downstream zero-shot transfer, rather than HPE feature reuse alone.

\fi
\section{PoseBridge}
\label{sec:method}

\subsection{Problem Setup, Recognition Protocol, and Overview}
\label{sec:overview}
\ifversionA
\else
Given an input video $\mathbf{V}=\{\mathbf{I}_t\}_{t=1}^{T}$, a pose estimator $\Phi_\mathrm{HPE}$ predicts joint coordinates $s_t=\Phi_\mathrm{HPE}(\mathbf{I}_t)$, forming the skeleton sequence $\mathbf{S}_i=\{s_t\}_{t=1}^{T}$ for the $i$-th sample. ZSSAR is trained on seen-class samples $\mathcal{D}^s=\{(\mathbf{S}_i,y_i)\}_{i=1}^{N_s}$ with $y_i\in\mathcal{Y}^s$ and evaluated on disjoint unseen classes $\mathcal{Y}^u$, where $\mathcal{Y}^s\cap\mathcal{Y}^u=\varnothing$. In ZSL, predictions are restricted to $\mathcal{Y}^u$; in GZSL, they are made over $\mathcal{Y}^s\cup\mathcal{Y}^u$.

\textbf{Recognition protocol}.\;
PoseBridge follows an HPE-aware ZSSAR protocol. Raw frames are accessed only through the same HPE process used to obtain skeletons. During ZSSAR training and inference, the HPE model is frozen and provides both skeleton sequence $\mathbf{S}_i$ and cached intermediate cue sequences $\mathbf{P}_i=\{\mathbf{p}_{i,t}\}_{t=1}^{T}$, where $\mathbf{p}_{i,t}$ denotes a frame-level pose-anchored cue. PoseBridge uses $\mathbf{S}_i$, $\mathbf{P}_i$, and text descriptions, explicitly leveraging RGB-derived HPE evidence without introducing a separate RGB action-recognition backbone or optical-flow branch. No unseen-class visual samples are used for training or prototype construction.

Typical skeletonize-then-align methods encode $\mathbf{S}_i$ as $\mathbf{z}_s=\mathcal{E}_s(\mathbf{S}_i)$ and align it with text prototypes $\mathbf{t}_c=\mathcal{E}_d(\mathbf{d}_c)$, where $\mathcal{E}_s$ and $\mathcal{E}_d$ denote the pretrained skeleton and text encoders. This alignment occurs after visual evidence has been compressed into joint coordinates. PoseBridge instead learns pose-anchored cues from intermediate HPE representations (Sec.~\ref{sec:method_HPE}) and transfers them to skeleton representations and prototypes (Sec.~\ref{sec:method_ZSSAR}); training and inference are described in Sec.~\ref{sec:training_inference}.
\fi

\begin{wrapfigure}{r}{0.48\textwidth}
\vspace{-15pt}
\centering
\includegraphics[width=0.48\textwidth]{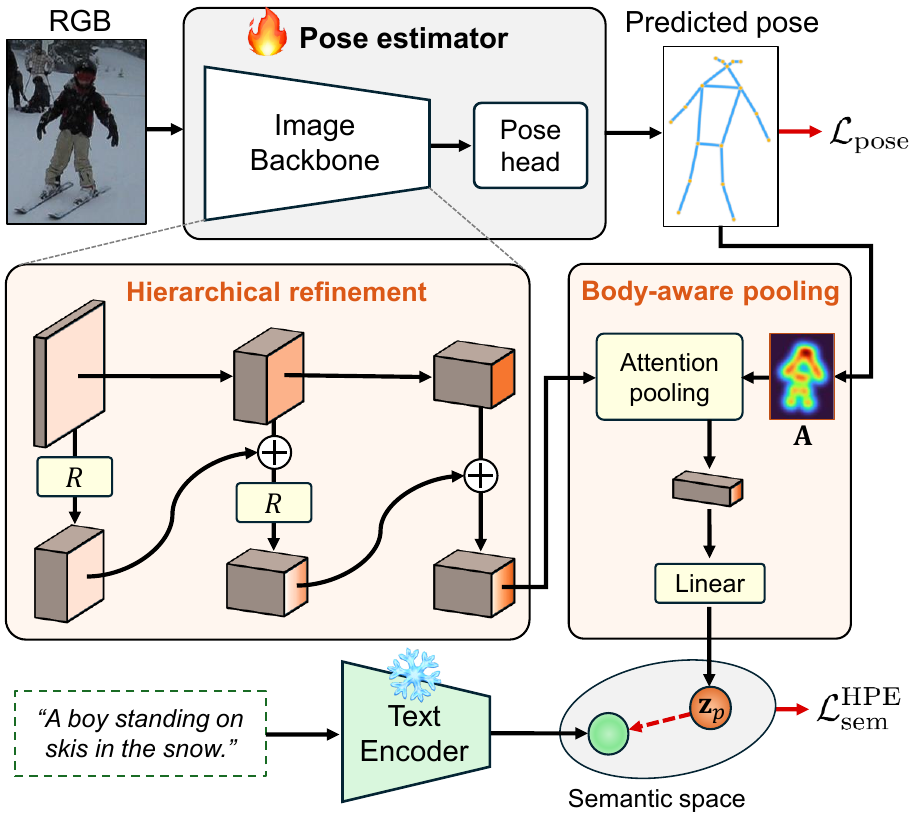}
\caption{Overview of PoseBridge for learning pose-anchored semantics within HPE.}
\label{fig:figure_HPE}
\vspace{-15pt}
\end{wrapfigure}

\subsection{Learning Pose-Anchored Semantics within HPE}
\label{sec:method_HPE}
HPE representations contain action-relevant visual evidence that is often lost after skeleton extraction. To preserve it, we convert intermediate HPE features into pose-anchored semantics, as illustrated in Fig.~\ref{fig:figure_HPE}. We refine multi-level features, aggregate them around the predicted body structure, and align them with action text semantics while preserving the original pose objective.

\textbf{Hierarchical pose-structured refinement}.\; Intermediate HPE layers provide complementary cues: shallow features retain local appearance and fine-grained spatial details, while deep features encode body configuration and keypoint structure. Using only the deepest feature can therefore discard visual evidence useful for actions requiring fine-grained visual distinctions. We address this by injecting shallower evidence into deeper pose features before semantic pooling.

Let $\{\mathbf{F}^l\}_{l=1}^{L}$ denote shallow-to-deep HPE features of frame $\mathbf{I}$, with $L=3$ in our experiments. Starting from $\tilde{\mathbf{F}}^1=\mathbf{F}^1$, each deeper feature is updated as:
\begin{equation}
    \tilde{\mathbf{F}}^{l+1} = \mathbf{F}^{l+1} + \alpha\,
    \mathcal{R}^l\big(\psi^l(\mathcal{P}^l(\tilde{\mathbf{F}}^l))\big),
    \qquad l=1, ..., L-1,
\end{equation}
where $\mathcal{P}^l$ projects the channel dimension and $\psi^l$ resizes the feature to the next-level resolution. $\mathcal{R}^l$ is a convolutional spatial refinement block applied before residual injection, and we fix $\alpha=0.5$. The final feature $\tilde{\mathbf{F}}^L$ preserves fine-grained visual evidence while maintaining high-level pose structure.

\textbf{Body-aware semantic pooling}.\; 
The refined feature map $\tilde{\mathbf{F}}^L$ contains responses over the full spatial map, but useful action semantics should be selected around the actor. We therefore build a body attention map $\mathbf{A}$ from the predicted joint probabilities: joint probabilities are converted into joint heatmaps, averaged over joints, and normalized to form a body-aware spatial prior. The frame-level pose-anchored cue $\mathbf{p}$ is then obtained by body-aware weighted pooling:
\begin{equation}
\mathbf{p} = W_p \left(\frac{\sum_{h,w} \mathbf{A}(h,w)\,\tilde{\mathbf{F}}^L(h,w)}
{\sum_{h,w} \mathbf{A}(h,w)+\varepsilon}\right).
\end{equation}
Here, $h,w$ index spatial locations in $\tilde{\mathbf{F}}^L$, $W_p$ projects the pooled feature into the semantic space, and $\varepsilon$ is used for numerical stability. Body-aware pooling encourages $\mathbf{p}$ to emphasize visual evidence near the predicted body structure, including local appearance cues around action-relevant body parts.

\textbf{Action-semantic alignment}.\; 
To align pose-anchored semantics with language, we use image-level captions from MS COCO~\cite{lin2014microsoft} as weak image-language supervision. Each paired description $\mathbf{d}$ is encoded by a pretrained CLIP text encoder $\mathcal{E}_d$ and projected into the same embedding space, yielding $\mathbf{t} = W_d\mathcal{E}_d(\mathbf{d})$.
Given a batch of $B$ pose-text pairs, we optimize a CLIP-style symmetric contrastive loss:
\begin{equation}
\mathcal{L}_{\mathrm{sem}}^{\mathrm{HPE}}=-\frac{1}{2B}
\sum_{i=1}^{B}\left[\log\frac{\exp(\mathbf{p}_{i}^\top \mathbf{t}_{i}/\tau)}
{\sum_{j=1}^{B}\exp(\mathbf{p}_{i}^\top \mathbf{t}_{j}/\tau)} 
+ 
\log\frac{\exp(\mathbf{p}_{i}^{\top}\mathbf{t}_{i}/\tau)}
{\sum_{j=1}^{B}\exp(\mathbf{p}_{j}^{\top}
\mathbf{t}_{i}/\tau)}\right],
\end{equation}
where $\mathbf{p}_i$ and $\mathbf{t}_i$ are L2-normalized embeddings of $i$-th sample, and $\tau=0.07$. This contrastive objective aligns pose-anchored cues with textual semantics while separating different samples. The pose estimator is trained with both pose and semantic losses:
\begin{equation}
\mathcal{L}_{\mathrm{HPE}} = \mathcal{L}_{\mathrm{pose}} +
\lambda_{\mathrm{hpe}}\,\mathcal{L}_{\mathrm{sem}}^{\mathrm{HPE}},
\end{equation}
where $\mathcal{L}_{\mathrm{pose}}$ is the original pose estimation loss and $\lambda_{\mathrm{hpe}}$ balances the two objectives. After training, the frozen HPE model extracts skeleton sequences $\mathbf{S}_i$ and pose-anchored semantic sequences $\mathbf{P}_i=\{\mathbf{p}_{i,t}\}_{t=1}^{T}$, which are passed to the ZSSAR stage as bridge evidence.

\begin{figure*}[t]
\begin{center}
\includegraphics[width=0.98\textwidth]{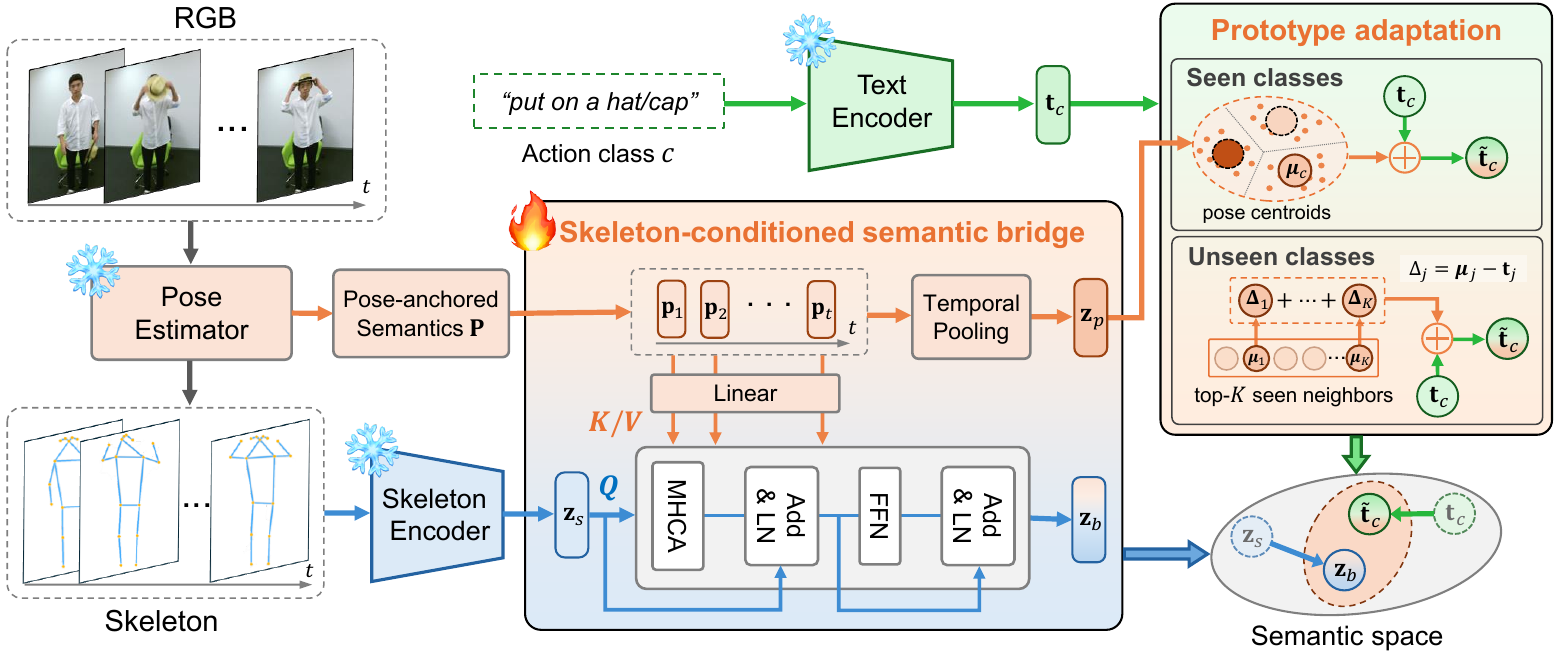}
\end{center}
\caption{Overview of PoseBridge in the ZSSAR. The pose-anchored semantics extracted from HPE are used for both the skeleton-conditioned semantic bridge and semantic prototype adaptation.}
\label{fig:figure_ZSSAR}
\end{figure*}

\subsection{Bridging Pose-Anchored Semantics to Skeleton-Text Alignment}
\label{sec:method_ZSSAR}
As illustrated in Fig.~\ref{fig:figure_ZSSAR}, we use the pose-anchored semantics learned within HPE to bridge skeleton representations and text prototypes in ZSSAR. The frozen HPE model extracts both the skeleton sequence $\mathbf{S}_i$ and the pose-anchored semantic sequence $\mathbf{P}_i$. PoseBridge uses $\mathbf{P}_i$ for two purposes: enriching the skeleton representation through a skeleton-conditioned semantic bridge and adapting text prototypes with pose-semantic cues. Thus, PoseBridge recovers action-relevant semantics from the skeleton extraction pipeline itself, without introducing an additional RGB recognition branch.

\textbf{Skeleton-conditioned semantic bridge}.\; 
Given the skeleton sequence $\mathbf{S}_i$, we first obtain the skeleton representation as $\mathbf{z}_s = W_s(\mathcal{E}_s(\mathbf{S}_i))$, where $\mathcal{E}_s$ is the skeleton encoder and $W_s$ projects the skeleton feature into the semantic embedding space. We then use $\mathbf{z}_s$ as a query and the temporal pose-anchored semantic sequence $\mathbf{P}_i$ as keys and values:
\begin{equation}
    \mathbf{z}_b = \mathrm{LN_2}(\mathbf{h}_i + \mathrm{FFN}(\mathbf{h}_i)),
    \qquad
    \mathbf{h}_i = \mathrm{LN_1}(\mathbf{z}_s + \sigma(\gamma_b) \odot \mathrm{MHCA}(\mathbf{z}_s, W_b\mathbf{P}_i)),
\end{equation}
where $W_b$ projects pose-anchored semantics into the bridge embedding space, $\mathrm{MHCA}$ denotes multi-head cross-attention, $\mathrm{LN}$ is layer normalization, $\sigma$ is the sigmoid function, and $\gamma_b$ is a learnable element-wise gate. This injects temporal pose-anchored semantics into the skeleton representation, producing $\mathbf{z}_b$ that remains grounded in skeleton motion while shifting toward a more semantically aligned space. Thus, the bridge makes complementary pose-anchored visual evidence accessible to skeleton-text matching, enabling the model to distinguish actions that are ambiguous from joint-coordinate motion alone.

\textbf{Semantic prototype adaptation}.\; 
Even after enriching the query with the skeleton-conditioned bridge, the target prototypes remain purely language-defined and may still be misaligned with the pose-semantic space. We therefore adapt text prototypes while keeping them as the primary anchors for zero-shot generality. From the temporal pose-anchored semantic sequence $\mathbf{P}_i$, we obtain a video-level pose-semantic representation $\mathbf{z}_{p,i}$ using attention-based temporal pooling. This converts frame-level HPE semantics into class-level pose evidence for prototype adaptation.

Let $\mathbf{t}_c$ denote the text prototype of class $c$. 
For each seen class, we compute the pose-semantic centroid from its training samples and use it for residual prototype adaptation:
\begin{equation}
    \tilde{\mathbf{t}}_c = (1-\rho)\mathbf{t}_c + \rho \bm\mu_c, 
    \qquad
    \bm\mu_c = \frac{1}{|\mathcal{D}^s_c|} \sum_{1 \le i\le|\mathcal{D}^s_c|} \mathbf{z}_{p,i}, \quad c \in \mathcal{Y}^s,
\end{equation}
where $\mathcal{D}^s_c$ denotes the training samples of seen class $c$, and $\rho$ is a fixed adaptation coefficient. This preserves the generality of the language prototype while correcting it toward the pose-semantic space.

For unseen classes, no pose samples are available. We therefore transfer the text-to-pose displacement from semantically related seen classes. Let $\mathcal{N}_K(c)$ denote the top-$K$ nearest seen classes to unseen class $c$ in the text prototype space. The unseen prototype is adapted as: 
\begin{equation}
    \tilde{\mathbf{t}}_{c} =
    \mathbf{t}_{c} + \rho \sum_{j \in \mathcal{N}_K(c)}
    \omega_{cj}(\bm\mu_j - \mathbf{t}_j),
    \qquad
    \omega_{cj} =
    \frac{
    \exp(\mathbf{t}_{c}^\top \mathbf{t}_j / \tau_a)
    }{
    \sum_{k \in \mathcal{N}_K(c)}
    \exp(\mathbf{t}_{c}^\top \mathbf{t}_k / \tau_a)
    },
    \quad c \in \mathcal{Y}^u.
\end{equation}
All centroids and adapted prototypes are L2-normalized. Since unseen adaptation uses only residual shifts from nearby seen classes without accessing unseen visual or pose samples, it preserves the zero-shot constraint while making text prototypes more compatible with the pose-semantic space.

\subsection{Training Objective and Inference}
\label{sec:training_inference}

\textbf{Training objective}.\;
ZSSAR is trained on seen classes only. We optimize three representations with different roles: the skeleton representation $\bf{z}_s$, the pose-semantic representation $\bf{z}_p$, and the bridge representation $\bf{z}_b$. 
While $\bf{z}_s$ and $\bf{z}_p$ serve as training anchors for shaping the motion and pose-semantic spaces, $\mathbf{z}_b$ is used as the final representation for zero-shot matching. Let $\mathbf{T}^s = [\mathbf{t}_c]_{c \in \mathcal{Y}^s}$ denote the seen-class text prototype matrix. We define three basic objectives:
\begin{equation}
\mathcal{L}_\mathrm{cls}(\mathbf{z}) = \mathrm{CE}(\mathcal{C}(\mathbf{z}), y),\quad
\mathcal{L}_\mathrm{sem}(\mathbf{z}) = \mathrm{CE}(\mathbf{z}\,(\mathbf{T}^s)^\top, y),\quad
\mathcal{L}_\mathrm{con}(\mathbf{z}) = \mathrm{SupCon}(\mathbf{z}, y),
\end{equation}
where $\mathcal{C}$ is a seen-class classifier, $\mathrm{CE}$ denotes cross-entropy loss, and $\mathrm{SupCon}$ denotes supervised contrastive loss \cite{khosla2020supervised}, which promotes intra-class compactness and inter-class separation.

For the skeleton and pose anchors, $r \in \{s, p\}$, we use all three losses: $\mathcal{L}_{r}=\lambda_{r}^\mathrm{cls}\mathcal{L}_\mathrm{cls}(\mathbf{z}_{r})+\lambda_{r}^\mathrm{sem}\mathcal{L}_\mathrm{sem}(\mathbf{z}_{r})+\lambda_{r}^\mathrm{con}\mathcal{L}_\mathrm{con}(\mathbf{z}_{r})$.
For the bridge representation, we omit the closed-set classifier and use only semantic matching and contrastive learning: $\mathcal{L}_{b}=\lambda_{b}^\mathrm{sem}\mathcal{L}_\mathrm{sem}(\mathbf{z}_{b})+\lambda_{b}^\mathrm{con}\mathcal{L}_\mathrm{con}(\mathbf{z}_{b})$.
This avoids forcing the inference representation $\mathbf{z}_b$ into seen-class classifier boundaries, keeping it better suited for prototype matching with unseen classes.
To couple the skeleton and pose-semantic representations, we further use cosine consistency and pose-guided distribution distillation:
\begin{equation}
\mathcal{L}_{\mathrm{align}}
=
\lambda_{\mathrm{s2p}}\left(1-\cos(\mathbf{z}_s,\mathbf{z}_p)\right)
+
\lambda_{\mathrm{kd}}\,
\mathrm{KL}\left(
\mathrm{softmax}\Bigl(
\frac{\mathbf{z}_p(\mathbf{T}^{s})^\top}{\tau_d}
\Bigr)^{\mathrm{sg}}
\middle\Vert
\mathrm{softmax}\Bigl(
\frac{\mathbf{z}_s(\mathbf{T}^{s})^\top}{\tau_d}
\Bigr)
\right),
\end{equation}
where $\mathrm{KL}$ denotes Kullback--Leibler divergence loss, $\tau_d$ is the distillation temperature, and $\mathrm{sg}$ denotes stop-gradient on the pose-guided teacher distribution. The final objective is $\mathcal{L}_\mathrm{ZSSAR} = \mathcal{L}_s + \mathcal{L}_p + \mathcal{L}_b + \mathcal{L}_\mathrm{align}$. Overall, this objective shapes $\mathbf{z}_s$ and $\mathbf{z}_p$ into discriminative semantic anchors, while training $\mathbf{z}_b$ as an open prototype-matching representation for unseen-class transfer.

\textbf{Inference}.\;
At inference, we use the bridge representation $\mathbf{z}_b$ as the final query and match it with the adapted prototypes $\tilde{\mathbf{t}}_c$. Since both are L2-normalized, similarity is computed by dot product, and ZSL/GZSL predictions are computed as:
\begin{equation}
\hat{y}_\mathrm{zsl}=\arg\max_{c\in\mathcal{Y}^u}\mathbf{z}_b^{\!\top}\tilde{\mathbf{t}}_c,\qquad
\hat{y}_\mathrm{gzsl}=\arg\max_{c\in\mathcal{Y}^s\cup\mathcal{Y}^u}\bigl(\mathbf{z}_b^{\!\top}\tilde{\mathbf{t}}_c-\kappa\,\mathbf{1}[c\in\mathcal{Y}^s]\bigr),
\end{equation}
where $\kappa$ is a seen-class calibration coefficient that mitigates seen-class bias in GZSL, following prior work~\cite{Chen_2025_CVPR}. Thus, inference explicitly couples the two sides of our bridge: $\mathbf{z}_b$ carries pose-anchored semantics recovered from the skeleton extraction process, while $\tilde{\mathbf{t}}$ provides skeleton-compatible semantic targets, jointly reducing the skeleton-text semantic gap.

\section{Experiments}
\label{sec:experiments}

\subsection{Experimental Setup}
\label{sec:experimental_setup}

\textbf{Datasets}.\; We evaluate PoseBridge on NTU-RGB+D 60 \cite{Shahroudy_2016_CVPR}, NTU-RGB+D 120 \cite{liu2019ntu}, PKU-MMD \cite{liu2017pku}, and Kinetics-200/400 \cite{kay2017kinetics,Zhu_2024_CVPR}. NTU-RGB+D 60/120 and PKU-MMD are controlled RGB-D skeleton action benchmarks, while Kinetics-200/400 provides in-the-wild videos with diverse scenes and action contexts. For all datasets, we follow the standard seen/unseen class splits used in prior ZSSAR works \cite{9506179, zhou2023zero, Zhu_2024_CVPR}. Detailed dataset descriptions are provided in Appendix~\ref{app:dataset_detail}.

\textbf{Evaluation protocol}.\; For zero-shot learning (ZSL), we report top-1 accuracy on unseen classes. For generalized zero-shot learning (GZSL), the prediction space contains both seen and unseen classes. Following prior works, we report the seen-class accuracy $S$, the unseen-class accuracy $U$, and their harmonic mean $H$, which measures the balance between seen and unseen recognition. 

\textbf{Implementation details}.\;
We use RTMPose-s \cite{jiang2023rtmpose} as the 2D pose estimator with RTMDet-nano \cite{lyu2022rtmdet} for person detection in the top-down HPE pipeline. The HPE model is trained with the original pose estimation objective and our action-semantic alignment objective on the MS COCO dataset \cite{lin2014microsoft}, and is then frozen to extract both estimated 2D skeleton sequences and pose-anchored semantics used in the ZSSAR stage. 
Following common ZSSAR evaluation protocols for fair comparison, we adopt Shift-GCN~\cite{Cheng_2020_CVPR} as the skeleton encoder and use the pretrained CLIP text encoder~\cite{radford2021learning} to obtain semantic prototypes from action descriptions. All models are trained only on seen classes and evaluated on disjoint unseen classes. 
Detailed implementation settings are provided in Appendix~\ref{app:implementation_details}, and hyperparameter/design choice analyses are presented in Appendix~\ref{app:design_analysis}.

\begin{table*}[t]
\caption{Comparison on NTU-RGB+D 60/120 under standard ZSL/GZSL splits. \textcolor{orange}{$^{\bm{\dagger}}$} denotes methods using an additional RGB pathway with a pretrained CLIP visual encoder. The best results are highlighted in \best{red}, and the second-best results are \second{blue}.
}
\label{tab:table_basic_split}
\vspace{3pt}
\centering
\setlength{\tabcolsep}{4.2pt}
\renewcommand{\arraystretch}{1.15}

\resizebox{\textwidth}{!}{%
\begin{tabular}{lc!{\datasetsep}
cccc!{\splitsep}
cccc!{\datasetsep}
cccc!{\splitsep}
cccc}
\toprule

\multirow{4}{*}{\textbf{Method}} 
& \multirow{4}{*}{\textbf{Venue}}
& \multicolumn{8}{c!{\datasetsep}}{\textbf{NTU-RGB+D 60 (Xsub)}}
& \multicolumn{8}{c}{\textbf{NTU-RGB+D 120 (Xsub)}} \\

\cmidrule(lr){3-10}\cmidrule(lr){11-18}

& 
& \multicolumn{4}{c!{\splitsep}}{55/5 Split}
& \multicolumn{4}{c!{\datasetsep}}{48/12 Split}
& \multicolumn{4}{c!{\splitsep}}{110/10 Split}
& \multicolumn{4}{c}{96/24 Split} \\

\cmidrule(lr){3-6}
\cmidrule(lr){7-10}
\cmidrule(lr){11-14}
\cmidrule(lr){15-18}

&
& \multicolumn{1}{c}{ZSL}
& \multicolumn{3}{c!{\splitsep}}{GZSL}
& \multicolumn{1}{c}{ZSL}
& \multicolumn{3}{c!{\datasetsep}}{GZSL}
& \multicolumn{1}{c}{ZSL}
& \multicolumn{3}{c!{\splitsep}}{GZSL}
& \multicolumn{1}{c}{ZSL}
& \multicolumn{3}{c}{GZSL} \\

\cmidrule(lr){3-3}\cmidrule(lr){4-6}
\cmidrule(lr){7-7}\cmidrule(lr){8-10}
\cmidrule(lr){11-11}\cmidrule(lr){12-14}
\cmidrule(lr){15-15}\cmidrule(lr){16-18}

&
& \textit{Acc} & \textit{S} & \textit{U} & \textit{H}
& \textit{Acc} & \textit{S} & \textit{U} & \textit{H}
& \textit{Acc} & \textit{S} & \textit{U} & \textit{H}
& \textit{Acc} & \textit{S} & \textit{U} & \textit{H} \\

\midrule

ReViSE \cite{Tsai_2017_ICCV} & ICCV 2017
& 53.9 & 74.2 & 34.7 & 29.2
& 17.5 & 62.4 & 20.8 & 31.2
& 55.0 & 48.7 & 44.8 & 46.7
& 32.4 & 49.7 & 25.1 & 33.3 \\

JPoSE \cite{Wray_2019_ICCV} & ICCV 2019
& 64.8 & 64.4 & 50.3 & 56.5
& 28.8 & 60.5 & 20.6 & 30.8
& 51.9 & 47.7 & 46.4 & 47.0
& 32.4 & 38.6 & 22.8 & 28.7 \\

CADA-VAE \cite{Schonfeld_2019_CVPR_Workshops} & CVPR 2019
& 76.8 & 69.4 & 61.8 & 65.4
& 29.0 & 51.3 & 27.0 & 35.4
& 59.5 & 47.2 & 19.8 & 48.4
& 35.8 & 41.1 & 34.1 & 37.3 \\

SynSE \cite{9506179} & ICIP 2021
& 75.8 & 61.3 & 56.9 & 59.0 
& 33.3 & 52.2 & 27.9 & 36.3  
& 62.7 & 52.5 & 57.6 & 54.9  
& 38.7 & 56.4 & 32.2 & 41.0  \\

SMIE \cite{zhou2023zero} & ACMMM 2023
& 78.0 & - & - & -
& 40.2 & - & - & -
& 65.7 & - & - & -
& 45.3 & - & - & - \\

PURLS \cite{Zhu_2024_CVPR} & CVPR 2024
& 79.2 & - & - & -
& 41.0 & - & - & -
& 72.0 & - & - & -
& 52.0 & - & - & - \\

SA-DVAE \cite{li2024sa} & ECCV 2024
& 82.4 & 62.3 & 70.8 & 66.3 
& 41.4 & 50.2 & 36.9 & 42.6 
& 68.8 & 61.1 & 59.8 & 60.4  
& 46.1 & 58.8 & 35.8 & 44.5 \\

STAR \cite{chen2024fine} & ACMMM 2024
& 81.4 & 69.0 & 69.9 & 69.4 
& 45.1 & 62.7 & 37.0 & 46.6 
& 63.3 & 59.9 & 52.7 & 56.1 
& 44.3 & 51.2 & 36.9 & 42.9 \\

SCoPLe \cite{Zhu_2025_CVPR} & CVPR 2025
& 84.1 & 69.6 & 71.9 & 70.8
& 53.0 & 54.5 & 61.8 & 57.9 
& 74.5 & 63.5 & 61.1 & 62.3 
& 52.2 & 53.3 & 51.2 & 52.2 \\

Neuron \cite{Chen_2025_CVPR} & CVPR 2025
& \second{86.9} & 69.1 & 73.8 & 71.4 
& \second{62.7} & 61.6 & 56.8 & \second{59.1} 
& 71.5 & 67.6 & 59.5 & 63.3
& 57.1 & 67.5 & 44.4 & 53.6 \\

TDSM \cite{Do_2025_ICCV} & ICCV 2025
& 86.5 & - & - & -
& 56.0 & - & - & -
& 74.2 & - & - & -
& \second{65.1} & - & - & - \\

FS-VAE \cite{Wu_2025_ICCV} & ICCV 2025
& \second{86.9} & 77.0 & 74.5 & 75.7
& 57.2 & 56.2 & 48.6 & 52.1
& 74.4 & 59.2 & 67.9 & 63.3
& 62.5 & 57.8 & 51.9 & 54.7 \\

\arrayrulecolor{black!60}
\hdashline
\arrayrulecolor{black}

{BSZSL\textcolor{orange}{$^{\bm{\dagger}}$}} \cite{liu2025beyond} & {ESWA 2025}
& {83.0} & {85.1} & {72.0} & \second{78.0}
& {53.0} & {73.9} & {36.5} & {48.8}
& \second{77.7} & {80.8} & {71.4} & \best{75.8}
& {56.1} & {72.1} & {53.8} & \second{61.6} \\

{SKI-VLM\textcolor{orange}{$^{\bm{\dagger}}$}} \cite{sinha2025ski} & {AAAI 2025}
& {82.2} & {-} & {-} & {-}
& {52.0} & {-} & {-} & {-}
& {77.5} & {-} & {-} & {-}
& {59.3} & {-} & {-} & {-} \\


\rowcolor{softlavender}
\textbf{PoseBridge} & \textbf{This work}
& \best{88.8} & 82.0 & 84.0 & \best{83.0}
& \best{73.2} & 75.6 & 61.9 & \best{68.0}
& \best{80.3} & 71.3 & 65.0 & \second{68.0}
& \best{70.9} & 66.1 & 59.4 & \best{62.6} \\

\bottomrule
\end{tabular}%
}
\end{table*}

\begin{table*}[t]
\caption{Comparison on NTU-RGB+D 60/120 and PKU-MMD under random ZSL/GZSL splits.}
\label{tab:table_random_split}
\vspace{3pt}
\centering
\setlength{\tabcolsep}{6.0pt}
\renewcommand{\arraystretch}{1.15}

\resizebox{\textwidth}{!}{%
\begin{tabular}{lc!{\datasetsep}
c!{\datasetsep}
c!{\datasetsep}
c}
\toprule

\multirow{3}{*}{\textbf{Method}}
& \multirow{3}{*}{\textbf{Venue}}
& \makebox[4.2cm][c]{\textbf{NTU-RGB+D 60 (Xsub)}}
& \makebox[4.2cm][c]{\textbf{NTU-RGB+D 120 (Xsub)}}
& \makebox[4.0cm][c]{\textbf{PKU-MMD (Xsub)}} \\

\cmidrule(lr){3-3}
\cmidrule(lr){4-4}
\cmidrule(lr){5-5}

&
& \makebox[3.8cm][c]{55/5 Split}
& \makebox[3.8cm][c]{110/10 Split}
& \makebox[3.8cm][c]{46/5 Split} \\

\cmidrule(lr){3-3}
\cmidrule(lr){4-4}
\cmidrule(lr){5-5}

&
& \makebox[1.9cm][c]{ZSL}\makebox[1.9cm][c]{GZSL \textit{(H)}}
& \makebox[1.9cm][c]{ZSL}\makebox[1.9cm][c]{GZSL \textit{(H)}}
& \makebox[1.9cm][c]{ZSL}\makebox[1.9cm][c]{GZSL \textit{(H)}} \\

\midrule

ReViSE \cite{Tsai_2017_ICCV} & ICCV 2017
& \makebox[1.9cm][c]{60.9}\makebox[1.9cm][c]{60.3}
& \makebox[1.9cm][c]{44.9}\makebox[1.9cm][c]{40.3}
& \makebox[1.9cm][c]{59.3}\makebox[1.9cm][c]{49.8} \\

JPoSE \cite{Wray_2019_ICCV} & ICCV 2019
& \makebox[1.9cm][c]{59.4}\makebox[1.9cm][c]{60.1}
& \makebox[1.9cm][c]{46.7}\makebox[1.9cm][c]{43.7}
& \makebox[1.9cm][c]{57.2}\makebox[1.9cm][c]{51.6} \\

CADA-VAE \cite{Schonfeld_2019_CVPR_Workshops} & CVPR 2019
& \makebox[1.9cm][c]{61.8}\makebox[1.9cm][c]{66.4}
& \makebox[1.9cm][c]{45.2}\makebox[1.9cm][c]{45.6}
& \makebox[1.9cm][c]{60.7}\makebox[1.9cm][c]{45.8} \\

SynSE \cite{9506179} & ICIP 2021
& \makebox[1.9cm][c]{64.2}\makebox[1.9cm][c]{67.5}
& \makebox[1.9cm][c]{47.3}\makebox[1.9cm][c]{43.5}
& \makebox[1.9cm][c]{53.9}\makebox[1.9cm][c]{49.5} \\

SMIE \cite{zhou2023zero} & ACMMM 2023
& \makebox[1.9cm][c]{65.1}\makebox[1.9cm][c]{-}
& \makebox[1.9cm][c]{46.4}\makebox[1.9cm][c]{-}
& \makebox[1.9cm][c]{60.8}\makebox[1.9cm][c]{-} \\

SA-DVAE \cite{li2024sa} & ECCV 2024
& \makebox[1.9cm][c]{84.2}\makebox[1.9cm][c]{75.3}
& \makebox[1.9cm][c]{50.7}\makebox[1.9cm][c]{47.5}
& \makebox[1.9cm][c]{66.5}\makebox[1.9cm][c]{54.7} \\

SCoPLe \cite{Zhu_2025_CVPR} & CVPR 2025
& \makebox[1.9cm][c]{83.7}\makebox[1.9cm][c]{\second{77.7}}
& \makebox[1.9cm][c]{53.3}\makebox[1.9cm][c]{\second{54.1}}
& \makebox[1.9cm][c]{\second{71.4}}\makebox[1.9cm][c]{54.9} \\

TDSM \cite{Do_2025_ICCV} & ICCV 2025
& \makebox[1.9cm][c]{\second{88.9}}\makebox[1.9cm][c]{-}
& \makebox[1.9cm][c]{\second{69.5}}\makebox[1.9cm][c]{-}
& \makebox[1.9cm][c]{70.8}\makebox[1.9cm][c]{-} \\

FS-VAE \cite{Wu_2025_ICCV} & ICCV 2025
& \makebox[1.9cm][c]{-}\makebox[1.9cm][c]{-}
& \makebox[1.9cm][c]{-}\makebox[1.9cm][c]{-}
& \makebox[1.9cm][c]{71.2}\makebox[1.9cm][c]{\second{59.0}} \\

\rowcolor{softlavender}
\textbf{PoseBridge} & \textbf{This work}
& \makebox[1.9cm][c]{\best{90.5}}\makebox[1.9cm][c]{\best{78.2}}
& \makebox[1.9cm][c]{\best{74.8}}\makebox[1.9cm][c]{\best{66.0}}
& \makebox[1.9cm][c]{\best{77.5}}\makebox[1.9cm][c]{\best{71.9}} \\

\bottomrule
\end{tabular}%
}
\end{table*}

\begin{table*}[t]
\caption{Results on Kinetics-200/400 under the PURLS \cite{Zhu_2024_CVPR} ZSL benchmark. $^*$ indicates the use of multiple text prompts; otherwise, a single text prompt is used.}
\label{tab:table_kinetics}
\vspace{3pt}
\centering
\setlength{\tabcolsep}{5.2pt}
\renewcommand{\arraystretch}{1.15}

\resizebox{\textwidth}{!}{%
\begin{tabular}{lc!{\datasetsep}
c!{\splitsep} c!{\splitsep} c!{\splitsep} c!{\datasetsep}
c!{\splitsep} c!{\splitsep} c!{\splitsep} c}
\toprule

\multirow{2}{*}{\textbf{Method}}
& \multirow{2}{*}{\textbf{Venue}}
& \multicolumn{4}{c!{\datasetsep}}{\textbf{Kinetics-200}}
& \multicolumn{4}{c}{\textbf{Kinetics-400}} \\

\cmidrule(lr){3-6}
\cmidrule(lr){7-10}

&
& 180/20 Split
& 160/40 Split
& 140/60 Split
& 120/80 Split
& 360/40 Split
& 320/80 Split
& 300/100 Split
& 280/120 Split \\

\midrule

DeVISE \cite{frome2013devise} & NeurIPS 2013
& 22.2 & 12.3 & 8.0 & 5.7
& 18.4 & 10.2 & 9.5 & 8.3 \\

ReViSE \cite{Tsai_2017_ICCV} & ICCV 2017
& 25.0 & 13.3 & 8.1 & 6.2
& 20.8 & 11.8 & 9.5 & 8.2 \\

PURLS \cite{Zhu_2024_CVPR} & CVPR 2024
& 26.0 & 15.9 & 10.2 & 7.8
& 22.5 & 15.1 & 11.4 & 11.0 \\

PURLS$^*$ \cite{Zhu_2024_CVPR} & CVPR 2024
& 32.2 & 22.6 & 12.0 & 11.8
& 34.5 & 24.3 & 17.0 & 14.3 \\

TDSM \cite{Do_2025_ICCV} & ICCV 2025
& \second{38.2} & \second{24.4} & \second{15.3} & \second{13.1}
& \second{38.9} & \second{26.2} & \second{18.5} & \second{16.1} \\

\rowcolor{softlavender}
\textbf{PoseBridge} & \textbf{This work}
& \best{55.6} & \best{38.3} & \best{28.6} & \best{26.4}
& \best{52.3} & \best{39.6} & \best{32.9} & \best{32.1} \\

\bottomrule
\end{tabular}%
}
\end{table*}

\subsection{Comparison with State-of-the-Art}

\textbf{Evaluation on standard split benchmark}.\; 
We compare PoseBridge with state-of-the-art ZSSAR methods on the standard NTU-RGB+D 60/120 Xsub protocols, following~\cite{9506179}. As shown in Table~\ref{tab:table_basic_split}, PoseBridge achieves the best ZSL performance across all settings and substantially improves GZSL performance. The gains are especially clear under harder splits with fewer seen and more unseen classes: on NTU-RGB+D 60 under the 48/12 split, PoseBridge improves the previous best ZSL accuracy by 10.5 points and the GZSL harmonic mean by 8.9 points. 
The strong results on NTU-RGB+D 120 under the 96/24 split further confirm the effectiveness of pose-anchored semantics for semantically diverse unseen actions.

Notably, PoseBridge outperforms recent RGB-assisted methods in almost all settings, despite using no additional RGB recognition branch. Instead, it recovers action-relevant visual semantics from intermediate HPE representations already computed during skeleton extraction, showing that the gain comes from better exploiting the skeletonization process rather than adding heavy RGB-based modules. To address potential concerns about skeleton input formats, Appendix~\ref{app:additional_evaluation_settings} provides a controlled comparison using the same RTMPose-extracted 2D skeletons, where PoseBridge remains superior, along with additional Xview and Xset results.

\textbf{Evaluation on random split benchmark}.\; 
Table~\ref{tab:table_random_split} reports results under the random split protocol introduced by SMIE~\cite{zhou2023zero}, where three randomly selected seen/unseen splits are evaluated and averaged. PoseBridge again delivers consistent improvements across NTU-RGB+D 60, NTU-RGB+D 120, and PKU-MMD datasets. These results demonstrate that the effectiveness of PoseBridge is consistent across different seen-unseen partitions, indicating that the proposed semantic bridge improves ZSSAR beyond a specific split configuration.

\textbf{Evaluation on Kinetics-200/400 dataset}.\; 
We further evaluate PoseBridge on the more challenging Kinetics-200/400 benchmarks following PURLS~\cite{Zhu_2024_CVPR}. Unlike the controlled NTU-RGB+D and PKU-MMD benchmarks, Kinetics contains in-the-wild videos with diverse scenes, viewpoints, subjects, and action contexts. As shown in Table~\ref{tab:table_kinetics}, PoseBridge substantially outperforms prior methods across all splits. In particular, PoseBridge improves Kinetics-200 under the 180/20 split by 17.4 points, from 38.2\% to 55.6\%, and Kinetics-400 under the 280/120 split by 16.0 points, from 16.1\% to 32.1\%. These substantial gains demonstrate the effectiveness of preserving pose-anchored visual evidence for zero-shot recognition, particularly when joint-coordinate motion alone is insufficient to distinguish actions.

\begin{table*}[t]
\caption{Accuracy-efficiency comparison on NTU-RGB+D 60/120 at the ZSSAR recognition stage. 
We report ZSL accuracy under standard splits, with parameters, GFLOPs, and throughput.}
\label{tab:efficiency_comparison}
\vspace{3pt}
\centering
\setlength{\tabcolsep}{5.2pt}
\renewcommand{\arraystretch}{1.15}

\resizebox{\textwidth}{!}{%
\begin{tabular}{lc!{\datasetsep}
c!{\splitsep} c!{\datasetsep}
c!{\splitsep} c!{\datasetsep}
c!{\splitsep}
c!{\splitsep}
c}
\toprule

\multirow{2}{*}{\textbf{Method}}
& \multirow{2}{*}{\textbf{Venue}}
& \multicolumn{2}{c!{\datasetsep}}{\textbf{NTU-RGB+D 60 (Xsub)}}
& \multicolumn{2}{c!{\datasetsep}}{\textbf{NTU-RGB+D 120 (Xsub)}}
& \multirow{2}{*}{\shortstack{\textbf{Trainable / Total}\\\textbf{\#Params (M)}}}
& \multirow{2}{*}{\shortstack{\textbf{GFLOPs}\\\textbf{(G)}}}
& \multirow{2}{*}{\shortstack{\textbf{Throughput}\\\textbf{(clips/s)}}} \\

\cmidrule(lr){3-4}
\cmidrule(lr){5-6}

&
&
55/5 Split
&
48/12 Split
&
110/10 Split
&
96/24 Split
&
&
&
\\

\midrule

Neuron~\cite{Chen_2025_CVPR} & CVPR 2025
& \second{86.9} & \second{62.7} & 71.5 & 57.1
& \second{2.3 / 126.7} & 7.5 & 1165.4 \\

TDSM~\cite{Do_2025_ICCV} & ICCV 2025
& 86.5 & 56.0 & 74.2 & \second{65.1}
& 260.4 / 384.2 & 10.6 & 772.5 \\

FS-VAE~\cite{Wu_2025_ICCV} & ICCV 2025
& \second{86.9} & 57.2 & 74.4 & 62.5
& \best{0.2 / 64.4} & \best{3.7} & \second{1237.0} \\

SKI-VLM~\cite{sinha2025ski} & AAAI 2025
& 82.2 & 52.0 & \second{77.5} & 59.3
& 2.8 / 215.9 & 281.0 & 22.6 \\

\rowcolor{softlavender}
\textbf{PoseBridge} & \textbf{This work}
& \best{88.8} & \best{73.2} & \best{80.3} & \best{70.9}
& 3.8 / 128.2 & \second{7.3} & \best{1398.0} \\

\bottomrule
\end{tabular}%
}
\end{table*}


\textbf{Computational cost}.\;
PoseBridge achieves a strong accuracy-efficiency trade-off among recent ZSSAR  approaches. As shown in Table~\ref{tab:efficiency_comparison}, it achieves the highest ZSL accuracy across all four standard NTU-RGB+D splits while maintaining the fastest throughput under the recognition-stage evaluation protocol. This comparison measures the ZSSAR recognition stage after skeleton and cue extraction, rather than the complete video-to-action pipeline. For all methods, trainable/total parameters, GFLOPs, and throughput are measured with the skeleton encoder and text encoder included on a single NVIDIA RTX A6000 GPU. Compared with SKI-VLM~\cite{sinha2025ski}, which relies on an additional RGB visual encoder and incurs substantially larger GFLOPs, PoseBridge recovers action-relevant semantics from intermediate HPE representations with a much more compact recognition pipeline. TDSM has the largest parameter count due to its diffusion-based backbone and text-conditioning components. Overall, PoseBridge improves zero-shot recognition while preserving the efficiency of a skeleton-based pipeline. Component-wise costs for the full PoseBridge pipeline, from pose estimation to ZSSAR prediction, are provided in Appendix~\ref{app:component_cost}.

\begin{wraptable}[13]{r}{0.46\linewidth}
\vspace{-20pt}
\caption{Ablation study of PoseBridge components on NTU-RGB+D 60 (48/12 split).}
\label{tab:ablation_components}
\vspace{2pt}
\centering
\scriptsize
\setlength{\tabcolsep}{4.5pt}
\renewcommand{\arraystretch}{1.02}

\resizebox{\linewidth}{!}{%
\begin{tabular}{cc cc!{\datasetsep} cc}
\toprule

\multicolumn{2}{c}{\textbf{HPE}}
& \multicolumn{2}{c!{\datasetsep}}{\textbf{ZSSAR}}
& \multicolumn{2}{c}{\textbf{NTU-RGB+D 60 (Xsub, 48/12 Split)}} \\

\cmidrule(lr){1-2}
\cmidrule(lr){3-4}
\cmidrule(lr){5-6}

HR & BP & SB & PA
& \makebox[1.2cm][c]{ZSL}
& \makebox[1.2cm][c]{GZSL \textit{(H)}} \\

\midrule

- & - & - & - 
& \makebox[1.2cm][c]{54.7} 
& \makebox[1.2cm][c]{52.0} \\

- & - & - & \checkmark 
& \makebox[1.2cm][c]{56.6} 
& \makebox[1.2cm][c]{57.3} \\

- & - & \checkmark & - 
& \makebox[1.2cm][c]{60.4} 
& \makebox[1.2cm][c]{60.7} \\

- & - & \checkmark & \checkmark 
& \makebox[1.2cm][c]{61.4} 
& \makebox[1.2cm][c]{61.7} \\

- & \checkmark & \checkmark & \checkmark 
& \makebox[1.2cm][c]{64.0} 
& \makebox[1.2cm][c]{63.2} \\

\checkmark & - & \checkmark & \checkmark 
& \makebox[1.2cm][c]{64.8} 
& \makebox[1.2cm][c]{63.8} \\

\checkmark & \checkmark & - & \checkmark 
& \makebox[1.2cm][c]{66.9} 
& \makebox[1.2cm][c]{63.9} \\

\checkmark & \checkmark & \checkmark & - 
& \makebox[1.2cm][c]{71.0} 
& \makebox[1.2cm][c]{66.5} \\

\rowcolor{softlavender}
\checkmark & \checkmark & \checkmark & \checkmark
& \makebox[1.2cm][c]{\textbf{73.2}}
& \makebox[1.2cm][c]{\textbf{68.0}} \\

\bottomrule
\end{tabular}%
}
\vspace{-6pt}
\end{wraptable}

\subsection{Ablation Study}
\label{sec:ablation_study}
\textbf{Component ablation.}
We analyze the contribution of each PoseBridge component in Table~\ref{tab:ablation_components}. The baseline uses only skeleton sequences from the original pose estimator, without HPE intermediate features or pose-anchored semantics. On the ZSSAR side, the skeleton-conditioned semantic bridge (SB) improves performance by injecting temporal pose-anchored cues, while prototype adaptation (PA) further aligns text prototypes with the pose-semantic space. 
On the HPE side, hierarchical refinement (HR) and body-aware pooling (BP) provide complementary gains by preserving and selecting body-aware semantics. The full model performs best, confirming the benefit of combining HPE-side semantic preservation with ZSSAR-side semantic transfer.

\begin{table}[h]
\caption{Matched raw-video comparisons on NTU-RGB+D 60 (48/12) and Kinetics-200 (120/80). Except for the RGB-only variant, all methods use skeleton inputs. PoseBridge-HPE denotes our learned HPE cues, and PoseBridge-ZSSAR comprises semantic bridging and prototype adaptation. Efficiency is measured over 16-frame clips.}
\label{tab:matched_controls}
\centering
\small
\setlength{\tabcolsep}{3.5pt}
\renewcommand{\arraystretch}{1.12}

\begin{tabular}{cllccrrr}
\toprule
& \multicolumn{2}{c}{\textbf{Configuration}}
& \multicolumn{2}{c}{\textbf{ZSL (\%)}}
& \multicolumn{3}{c}{\textbf{End-to-end efficiency}} \\
\cmidrule(lr){2-3}
\cmidrule(lr){4-5}
\cmidrule(l){6-8}

\textbf{Variant}
& \textbf{Visual cue source}
& \textbf{Integration}
& \textbf{NTU60}
& \textbf{K200}
& \textbf{Params}
& \textbf{GFLOPs}
& \textbf{Throughput} \\
\midrule

A & None        & Skeleton-only
  & 54.7 & 19.1 & 132.0 & 23.47 & 238.6 \\
B & VideoCLIP   & RGB-only
  & 47.2 & 27.8 & 212.7 & 77.28 & 112.4 \\
C & VideoCLIP   & Simple fusion
  & 59.6 & \textbf{29.6} & 220.1 & 94.09 & 79.0 \\
D & Vanilla HPE & Simple fusion
  & 57.3 & 19.3 & 132.2 & 23.47 & 235.2 \\
E & PoseBridge-HPE      & Simple fusion
  & 62.3 & 22.3 & 135.6 & 27.92 & 233.0 \\
F & Vanilla HPE & PoseBridge-ZSSAR
  & 61.4 & 19.2 & 134.7 & 23.49 & 233.6 \\
\rowcolor{softlavender}
\textbf{G} & \textbf{PoseBridge-HPE} & \textbf{PoseBridge-ZSSAR}
  & \textbf{73.2} & 26.4 & 138.1 & 27.94 & 232.8 \\
\bottomrule
\end{tabular}
\end{table}

\noindent\textbf{Matched controls.}
To isolate the effects of visual cues and integration strategies, Table~\ref{tab:matched_controls} compares skeleton-only recognition (A), large-scale pretrained VideoCLIP \cite{xu2021videoclip} as a separate RGB branch (B), and its fusion with skeleton features (C), alongside HPE-based variants (D--G). Vanilla HPE cues reuse intermediate features learned solely for pose estimation, whereas PoseBridge-HPE explicitly learns pose-anchored semantics. Simple fusion concatenates skeleton and visual features followed by a two-layer MLP. Under identical fusion, PoseBridge-HPE improves NTU60/K200 accuracy by 5.0/3.0 points (D to E), while PB-ZSSAR further improves it to 73.2/26.4 (E to G), demonstrating complementary semantic preservation and transfer. Compared with VideoCLIP fusion (C), PoseBridge achieves higher NTU60 accuracy (73.2 vs. 59.6), though lower on K200 (26.4 vs. 29.6), while requiring 3.4$\times$ fewer GFLOPs and achieving 2.9$\times$ higher throughput without a separate RGB backbone.

\subsection{Qualitative Analysis}
\label{sec:qualitative}
\begin{figure*}[t]
\begin{center}
\includegraphics[width=1.0\textwidth]{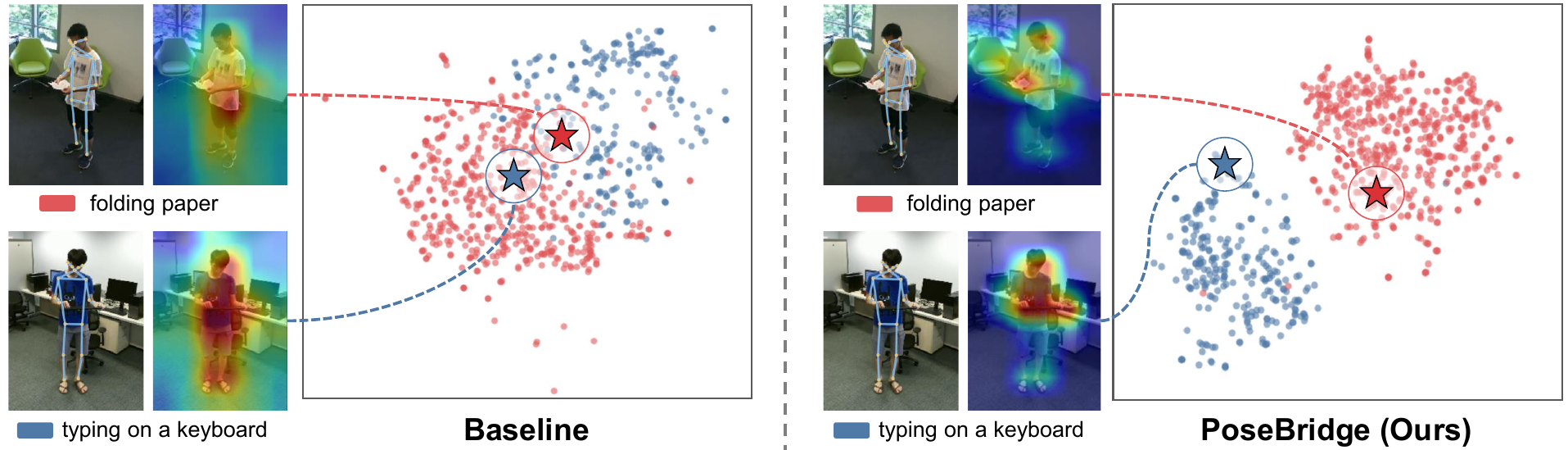}
\end{center}
\caption{Qualitative comparison between the baseline and PoseBridge on the confusing action pair ``folding paper'' and ``typing on a keyboard''. We visualize predicted skeletons, HPE Grad-CAM maps, and t-SNE embeddings of zero-shot matching features.}
\label{fig:qualitative_analysis}
\end{figure*}

We qualitatively verify that PoseBridge recovers semantic cues that are largely missing from skeleton coordinates alone. In Fig.~\ref{fig:qualitative_analysis}, we compare a baseline trained only with the original pose estimation objective and our HPE model trained with action-semantic alignment, hierarchical refinement, and body-aware pooling. For the confusing action pair ``folding paper'' and ``typing on a keyboard'', Grad-CAM shows that the baseline mainly responds to the overall human body, reflecting its bias toward keypoint localization. In contrast, our pose-anchored feature focuses more strongly on action-relevant regions, such as the keyboard, the paper, and hand-object interaction areas. The t-SNE plots show a consistent trend: baseline features are highly mixed, while PoseBridge yields much clearer separation between the two classes. These results show that PoseBridge captures action-relevant semantics during HPE and transfers them effectively to skeleton-text matching in zero-shot recognition.

\section{Conclusion}
\label{sec:conclusion}

We presented PoseBridge, an HPE-aware framework for ZSSAR that addresses a fundamental limitation of skeletonize-then-align approaches: the loss of action-relevant visual evidence during skeletonization. Rather than treating HPE solely as preprocessing, PoseBridge preserves pose-anchored semantics from intermediate HPE representations and transfers them to skeleton-text alignment through a skeleton-conditioned semantic bridge and prototype adaptation. Experiments on NTU-RGB+D 60/120, PKU-MMD, and Kinetics-200/400 demonstrate consistent improvements, particularly on challenging splits and in-the-wild videos. Importantly, PoseBridge achieves these gains without a separate RGB action-recognition backbone, highlighting HPE as an effective upstream semantic source. Overall, PoseBridge demonstrates the importance of preserving visual semantics before skeleton-text alignment for advancing zero-shot recognition.

\begin{ack}
This work was supported by Artificial intelligence industrial convergence cluster development project funded by the Ministry of Science and ICT (MSIT, Korea)\&Gwangju Metropolitan City, and in part by the Basic Science Research Program through the National Research Foundation of Korea (NRF) funded by the Ministry of Education (RS-2025-25425223), and in part by Basic Science Research Program through the National Research Foundation of Korea (NRF) funded by the Ministry of Education (RS-2025-25416432).
\end{ack}

\bibliographystyle{plainnat}
\bibliography{ref}

@InProceedings{Tsai_2017_ICCV,
title={Learning robust visual-semantic embeddings},
  author={Hubert Tsai, Yao-Hung and Huang, Liang-Kang and Salakhutdinov, Ruslan},
  booktitle={Proceedings of the IEEE International conference on Computer Vision},
  pages={3571--3580},
  year={2017}
}

@InProceedings{Wray_2019_ICCV,
  title={Fine-grained action retrieval through multiple parts-of-speech embeddings},
  author={Wray, Michael and Larlus, Diane and Csurka, Gabriela and Damen, Dima},
  booktitle={Proceedings of the IEEE/CVF international conference on computer vision},
  pages={450--459},
  year={2019}
}

@InProceedings{Schonfeld_2019_CVPR_Workshops,
  title={Generalized zero-shot learning via aligned variational autoencoders},
  author={Schonfeld, Edgar and Ebrahimi, Sayna and Sinha, Samarth and Darrell, Trevor and Akata, Zeynep},
  booktitle={Proceedings of the IEEE/CVF Conference on Computer Vision and Pattern Recognition Workshops},
  pages={54--57},
  year={2019}
}

@inproceedings{9506179,
  title={Syntactically guided generative embeddings for zero-shot skeleton action recognition},
  author={Gupta, Pranay and Sharma, Divyanshu and Sarvadevabhatla, Ravi Kiran},
  booktitle={2021 IEEE International Conference on Image Processing (ICIP)},
  pages={439--443},
  year={2021},
  organization={IEEE}
}

@inproceedings{zhou2023zero,
  title={Zero-shot skeleton-based action recognition via mutual information estimation and maximization},
  author={Zhou, Yujie and Qiang, Wenwen and Rao, Anyi and Lin, Ning and Su, Bing and Wang, Jiaqi},
  booktitle={Proceedings of the 31st ACM international conference on multimedia},
  pages={5302--5310},
  year={2023}
}

@InProceedings{Zhu_2024_CVPR,
  title={Part-aware unified representation of language and skeleton for zero-shot action recognition},
  author={Zhu, Anqi and Ke, Qiuhong and Gong, Mingming and Bailey, James},
  booktitle={Proceedings of the IEEE/CVF Conference on Computer Vision and Pattern Recognition},
  pages={18761--18770},
  year={2024}
}

@inproceedings{li2024sa,
  title={Sa-dvae: Improving zero-shot skeleton-based action recognition by disentangled variational autoencoders},
  author={Li, Sheng-Wei and Wei, Zi-Xiang and Chen, Wei-Jie and Yu, Yi-Hsin and Yang, Chih-Yuan and Hsu, Jane Yung-jen},
  booktitle={European conference on computer vision},
  pages={447--462},
  year={2024},
  organization={Springer}
}

@inproceedings{chen2024fine,
  title={Fine-grained side information guided dual-prompts for zero-shot skeleton action recognition},
  author={Chen, Yang and Guo, Jingcai and He, Tian and Lu, Xiaocheng and Wang, Ling},
  booktitle={Proceedings of the 32nd ACM International Conference on Multimedia},
  pages={778--786},
  year={2024}
}

@InProceedings{Zhu_2025_CVPR,
  title={Semantic-guided cross-modal prompt learning for skeleton-based zero-shot action recognition},
  author={Zhu, Anqi and Zhu, Jingmin and Bailey, James and Gong, Mingming and Ke, Qiuhong},
  booktitle={Proceedings of the IEEE/CVF Conference on Computer Vision and Pattern Recognition},
  pages={13876--13885},
  year={2025}
}

@InProceedings{Chen_2025_CVPR,
  title={Neuron: Learning context-aware evolving representations for zero-shot skeleton action recognition},
  author={Chen, Yang and Guo, Jingcai and Guo, Song and Tao, Dacheng},
  booktitle={Proceedings of the Computer Vision and Pattern Recognition Conference},
  pages={8721--8730},
  year={2025}
}

@InProceedings{Wu_2025_ICCV,
  title={Frequency-semantic enhanced variational autoencoder for zero-shot skeleton-based action recognition},
  author={Wu, Wenhan and Guo, Zhishuai and Chen, Chen and Xue, Hongfei and Lu, Aidong},
  booktitle={Proceedings of the IEEE/CVF International Conference on Computer Vision},
  pages={11122--11131},
  year={2025}
}

@InProceedings{Do_2025_ICCV,
    author    = {Do, Jeonghyeok and Kim, Munchurl},
    title     = {Bridging the Skeleton-Text Modality Gap: Diffusion-Powered Modality Alignment for Zero-shot Skeleton-based Action Recognition},
    booktitle = {Proceedings of the IEEE/CVF International Conference on Computer Vision (ICCV)},
    month     = {October},
    year      = {2025},
    pages     = {12757-12768}
}

@article{liu2025beyond,
  title={Beyond-Skeleton: Zero-shot Skeleton Action Recognition enhanced by supplementary RGB visual information},
  author={Liu, Hongjie and Niu, Yingchun and Zeng, Kun and Liu, Chun and Hu, Mengjie and Song, Qing},
  journal={Expert Systems with Applications},
  volume={273},
  pages={126814},
  year={2025},
  publisher={Elsevier}
}

@inproceedings{sinha2025ski,
  title={Ski models: Skeleton induced vision-language embeddings for understanding activities of daily living},
  author={Sinha, Arkaprava and Reilly, Dominick and Bremond, Francois and Wang, Pu and Das, Srijan},
  booktitle={Proceedings of the AAAI Conference on Artificial Intelligence},
  volume={39},
  number={7},
  pages={6931--6939},
  year={2025}
}

@article{frome2013devise,
  title={Devise: A deep visual-semantic embedding model},
  author={Frome, Andrea and Corrado, Greg S and Shlens, Jon and Bengio, Samy and Dean, Jeff and Ranzato, Marc'Aurelio and Mikolov, Tomas},
  journal={Advances in neural information processing systems},
  volume={26},
  year={2013}
}

@article{jiang2023rtmpose,
  title={Rtmpose: Real-time multi-person pose estimation based on mmpose},
  author={Jiang, Tao and Lu, Peng and Zhang, Li and Ma, Ningsheng and Han, Rui and Lyu, Chengqi and Li, Yining and Chen, Kai},
  journal={arXiv preprint arXiv:2303.07399},
  year={2023}
}

@inproceedings{lin2014microsoft,
  title={Microsoft coco: Common objects in context},
  author={Lin, Tsung-Yi and Maire, Michael and Belongie, Serge and Hays, James and Perona, Pietro and Ramanan, Deva and Doll{\'a}r, Piotr and Zitnick, C Lawrence},
  booktitle={European conference on computer vision},
  pages={740--755},
  year={2014},
  organization={Springer}
}

@InProceedings{Shahroudy_2016_CVPR,
  title={Ntu rgb+ d: A large scale dataset for 3d human activity analysis},
  author={Shahroudy, Amir and Liu, Jun and Ng, Tian-Tsong and Wang, Gang},
  booktitle={Proceedings of the IEEE conference on computer vision and pattern recognition},
  pages={1010--1019},
  year={2016}
}

@article{liu2019ntu,
  title={Ntu rgb+ d 120: A large-scale benchmark for 3d human activity understanding},
  author={Liu, Jun and Shahroudy, Amir and Perez, Mauricio and Wang, Gang and Duan, Ling-Yu and Kot, Alex C},
  journal={IEEE transactions on pattern analysis and machine intelligence},
  volume={42},
  number={10},
  pages={2684--2701},
  year={2019},
  publisher={IEEE}
}

@article{liu2017pku,
  title={Pku-mmd: A large scale benchmark for continuous multi-modal human action understanding},
  author={Liu, Chunhui and Hu, Yueyu and Li, Yanghao and Song, Sijie and Liu, Jiaying},
  journal={arXiv preprint arXiv:1703.07475},
  year={2017}
}

@article{kay2017kinetics,
  title={The kinetics human action video dataset},
  author={Kay, Will and Carreira, Joao and Simonyan, Karen and Zhang, Brian and Hillier, Chloe and Vijayanarasimhan, Sudheendra and Viola, Fabio and Green, Tim and Back, Trevor and Natsev, Paul and others},
  journal={arXiv preprint arXiv:1705.06950},
  year={2017}
}

@article{khosla2020supervised,
  title={Supervised contrastive learning},
  author={Khosla, Prannay and Teterwak, Piotr and Wang, Chen and Sarna, Aaron and Tian, Yonglong and Isola, Phillip and Maschinot, Aaron and Liu, Ce and Krishnan, Dilip},
  journal={Advances in neural information processing systems},
  volume={33},
  pages={18661--18673},
  year={2020}
}

@article{wang2026skeletoncontext,
  title={SkeletonContext: Skeleton-side Context Prompt Learning for Zero-Shot Skeleton-based Action Recognition},
  author={Wang, Ning and Wu, Tieyue and Sharif, Naeha and Boussaid, Farid and Zhu, Guangming and Mei, Lin and Bennamoun, Mohammed and others},
  journal={arXiv preprint arXiv:2603.29692},
  year={2026}
}

@article{jasani2019skeleton,
  title={Skeleton based zero shot action recognition in joint pose-language semantic space},
  author={Jasani, Bhavan and Mazagonwalla, Afshaan},
  journal={arXiv preprint arXiv:1911.11344},
  year={2019}
}

@article{chen2025learning,
  title={Learning by Neighbor-Aware Semantics, Deciding by Open-form Flows: Towards Robust Zero-Shot Skeleton Action Recognition},
  author={Chen, Yang and Li, Miaoge and Rao, Zhijie and Zeng, Deze and Guo, Song and Guo, Jingcai},
  journal={arXiv preprint arXiv:2511.09388},
  year={2025}
}

@article{wen2025enhancing,
  title={Enhancing skeleton-based action recognition with feature maps from pose estimation networks},
  author={Wen, Hao and Lu, Zhe-Ming and Shen, Fengli and Lu, Ziqian and Zheng, Yangming and Cui, Jialin},
  journal={IEICE Transactions on Fundamentals of Electronics, Communications and Computer Sciences},
  pages={2024EAP1162},
  year={2025},
  publisher={The Institute of Electronics, Information and Communication Engineers}
}

@InProceedings{Cheng_2020_CVPR,
  title={Skeleton-based action recognition with shift graph convolutional network},
  author={Cheng, Ke and Zhang, Yifan and He, Xiangyu and Chen, Weihan and Cheng, Jian and Lu, Hanqing},
  booktitle={Proceedings of the IEEE/CVF conference on computer vision and pattern recognition},
  pages={183--192},
  year={2020}
}

@inproceedings{radford2021learning,
  title={Learning transferable visual models from natural language supervision},
  author={Radford, Alec and Kim, Jong Wook and Hallacy, Chris and Ramesh, Aditya and Goh, Gabriel and Agarwal, Sandhini and Sastry, Girish and Askell, Amanda and Mishkin, Pamela and Clark, Jack and others},
  booktitle={International conference on machine learning},
  pages={8748--8763},
  year={2021},
  organization={PmLR}
}

@inproceedings{sun2019deep,
  title={Deep high-resolution representation learning for human pose estimation},
  author={Sun, Ke and Xiao, Bin and Liu, Dong and Wang, Jingdong},
  booktitle={Proceedings of the IEEE/CVF conference on computer vision and pattern recognition},
  pages={5693--5703},
  year={2019}
}

@article{xu2022vitpose,
  title={Vitpose: Simple vision transformer baselines for human pose estimation},
  author={Xu, Yufei and Zhang, Jing and Zhang, Qiming and Tao, Dacheng},
  journal={Advances in neural information processing systems},
  volume={35},
  pages={38571--38584},
  year={2022}
}

@article{khanam2024yolov11,
  title={Yolov11: An overview of the key architectural enhancements},
  author={Khanam, Rahima and Hussain, Muhammad},
  journal={arXiv preprint arXiv:2410.17725},
  year={2024}
}

@inproceedings{liu2018recognizing,
  title={Recognizing human actions as the evolution of pose estimation maps},
  author={Liu, Mengyuan and Yuan, Junsong},
  booktitle={Proceedings of the IEEE conference on computer vision and pattern recognition},
  pages={1159--1168},
  year={2018}
}

@inproceedings{duan2022revisiting,
  title={Revisiting skeleton-based action recognition},
  author={Duan, Haodong and Zhao, Yue and Chen, Kai and Lin, Dahua and Dai, Bo},
  booktitle={Proceedings of the IEEE/CVF conference on computer vision and pattern recognition},
  pages={2969--2978},
  year={2022}
}

@inproceedings{yan2018spatial,
  title={Spatial temporal graph convolutional networks for skeleton-based action recognition},
  author={Yan, Sijie and Xiong, Yuanjun and Lin, Dahua},
  booktitle={Proceedings of the AAAI conference on artificial intelligence},
  volume={32},
  number={1},
  year={2018}
}

@inproceedings{shi2019two,
  title={Two-stream adaptive graph convolutional networks for skeleton-based action recognition},
  author={Shi, Lei and Zhang, Yifan and Cheng, Jian and Lu, Hanqing},
  booktitle={Proceedings of the IEEE/CVF conference on computer vision and pattern recognition},
  pages={12026--12035},
  year={2019}
}

@article{kuang2025zero,
  title={Zero-shot skeleton-based action recognition with dual visual-text alignment},
  author={Kuang, Jidong and Wang, Hongsong and Han, Chaolei and Zhang, Yang and Gui, Jie},
  journal={Pattern Recognition},
  pages={112342},
  year={2025},
  publisher={Elsevier}
}

@article{lyu2022rtmdet,
  title={Rtmdet: An empirical study of designing real-time object detectors},
  author={Lyu, Chengqi and Zhang, Wenwei and Huang, Haian and Zhou, Yue and Wang, Yudong and Liu, Yanyi and Zhang, Shilong and Chen, Kai},
  journal={arXiv preprint arXiv:2212.07784},
  year={2022}
}

@inproceedings{xu2021videoclip,
  title={Videoclip: Contrastive pre-training for zero-shot video-text understanding},
  author={Xu, Hu and Ghosh, Gargi and Huang, Po-Yao and Okhonko, Dmytro and Aghajanyan, Armen and Metze, Florian and Zettlemoyer, Luke and Feichtenhofer, Christoph},
  booktitle={Proceedings of the 2021 conference on empirical methods in natural language processing},
  pages={6787--6800},
  year={2021}
}








\clearpage
\appendix

\section*{Appendix}

This appendix provides additional details, results, and analyses that support the main paper. 
Section~\ref{app:implementation_details} describes the implementation details of the HPE and ZSSAR stages. 
Section~\ref{app:dataset_detail} summarizes the datasets, seen/unseen split settings, and action descriptions. 
Section~\ref{app:additional_evaluation_settings} reports additional performance comparisons, including challenging splits, Xview/Xset protocols, controlled comparisons with estimated 2D skeleton inputs, and the effect of semantic alignment on pose estimation. 
Section~\ref{app:design_analysis} analyzes key design choices, including hyperparameters, component-wise computational costs, and different pose estimators. 
Section~\ref{app:additional_analyses} provides confusion matrix comparisons and additional analyses of object-sensitive recognition, representation quality, robustness, semantic supervision, and statistical reliability. 
Section~\ref{app:additional_qualitative_results} presents additional qualitative results, including t-SNE and Grad-CAM visualizations. 
Section~\ref{app:limitations_broader_impacts} discusses limitations and broader impacts.

\section{Implementation Details}
\label{app:implementation_details}

\subsection{HPE Details}
\label{app:hpe_details}
We use RTMPose-s~\cite{jiang2023rtmpose} as the 2D pose estimator and train it on MS COCO~\cite{lin2014microsoft}. For the top-down pose estimation pipeline, we employ RTMDet-nano \cite{lyu2022rtmdet} for person detection, followed by RTMPose-s for keypoint estimation. Both the skeleton coordinates and pose-anchored semantic cues are extracted from this shared HPE pipeline. Following the standard RTMPose-s setting, we use an input resolution of $256 \times 192$ and train the model for $420$ epochs with AdamW. The base learning rate is set to $4\times10^{-3}$, the weight decay is set to $0$, and the learning rate is adjusted with a linear warm-up followed by cosine annealing. The model is optimized with the original SimCC-based pose estimation loss and our action-semantic alignment loss described in Sec.~\ref{sec:method_HPE}.

For action-semantic alignment, we use COCO captions as image-level textual supervision. 
We encode each caption using a frozen CLIP ViT-B/32 text encoder~\cite{radford2021learning} and map the resulting feature to the pose-semantic embedding space with a lightweight projection head. 
The projection dimension is set to $512$, and the semantic loss weight is set to $\lambda_{\mathrm{hpe}}=0.1$. 
The CLIP-style contrastive loss for semantic alignment is applied from the beginning of training.

For the pose-anchored semantic cue extractor, we use $L=3$ feature levels by default. 
The hierarchical refinement gate is fixed to $\alpha=0.5$. After training, the HPE model is frozen and used to extract both 2D skeleton sequences and pose-anchored semantic cues for the ZSSAR stage. Additional analyses on design choices, including different pose estimators such as HRNet~\cite{sun2019deep} and ViTPose~\cite{xu2022vitpose}, are reported in Appendix~\ref{app:design_analysis}.

\subsection{ZSSAR Details}
\label{app:zssar_details}
We use Shift-GCN~\cite{Cheng_2020_CVPR} as the skeleton encoder and the pretrained CLIP ViT-L/14 text encoder~\cite{radford2021learning} for semantic prototype extraction. The CLIP text encoder is frozen, and action descriptions are encoded and cached before training. The skeleton feature dimension is $256$, the pose-semantic cue dimension is $512$, and the shared embedding dimension is $512$. The skeleton projector uses two MLP layers, while the pose adapter uses one MLP layer. The dropout ratio is set to $0.1$.

The skeleton-conditioned semantic bridge uses multi-head cross-attention with $4$ heads. 
For each video, we use $16$ temporal pose-anchored cues. The skeleton representation is used as the query, and the temporal pose-anchored cues are used as keys and values. For prototype adaptation, we use the top-$K=5$ nearest seen classes in the text-prototype space. 
The prototype adaptation coefficient is set to $\rho=0.2$.

We train PoseBridge with AdamW for $30$ epochs. 
The batch size is $128$, the test batch size is $256$, the learning rate is $1\times10^{-3}$, and the weight decay is $2\times10^{-3}$. 
We use $5$ warm-up epochs followed by cosine annealing with a minimum learning rate of $1\times10^{-6}$. 
Gradient clipping is applied with a maximum norm of $1.0$. 
We also use an exponential moving average (EMA) with decay $0.999$ starting from epoch $5$. All experiments are conducted on a single NVIDIA RTX A6000 GPU.

For the loss weights, we set 
$\lambda_s^{\mathrm{cls}}=1.0$, 
$\lambda_s^{\mathrm{sem}}=1.5$, 
$\lambda_s^{\mathrm{con}}=1.5$, 
$\lambda_p^{\mathrm{cls}}=0.5$, 
$\lambda_p^{\mathrm{sem}}=0.5$, 
$\lambda_p^{\mathrm{con}}=0.3$, 
$\lambda_{\mathrm{s2p}}=0.3$, 
$\lambda_{\mathrm{kd}}=1.0$, 
$\lambda_b^{\mathrm{sem}}=1.0$, and 
$\lambda_b^{\mathrm{con}}=0.5$. 
The supervised contrastive temperature is set to $0.07$, and the distillation temperature is set to $4.0$.
Additional analyses on design choices are reported in Appendix~\ref{app:design_analysis}.

\section{Dataset Details}
\label{app:dataset_detail}

\subsection{Dataset Details}

\textbf{NTU-RGB+D 60}.\;
NTU-RGB+D 60 \cite{Shahroudy_2016_CVPR} is a large-scale RGB-D action recognition benchmark containing 56,880 samples from 60 action classes performed by 40 subjects. 
The dataset was captured by three Kinect V2 cameras and provides multi-modal observations, including RGB videos, depth maps, infrared sequences, and 3D skeleton annotations. 
The standard evaluation protocols are cross-subject (Xsub) and cross-view (Xview), which evaluate generalization across different subjects and camera viewpoints, respectively. For zero-shot evaluation, we follow the seen/unseen class splits adopted in prior ZSSAR works \cite{9506179, zhou2023zero, Zhu_2024_CVPR}.

\textbf{NTU-RGB+D 120}.\;
NTU-RGB+D 120 \cite{liu2019ntu} extends NTU-RGB+D 60 to 114,480 samples from 120 action classes. It is also captured with Kinect V2 sensors and provides RGB, depth, infrared, and skeleton modalities. Compared with NTU-RGB+D 60, it contains a larger action vocabulary and more diverse acquisition setups, making it a more challenging benchmark for zero-shot generalization. The standard evaluation protocols are cross-subject (Xsub) and cross-setup (Xset), where Xset evaluates generalization across different camera and environment setups. We follow the standard ZSSAR seen/unseen splits used in previous works \cite{9506179, zhou2023zero, Zhu_2024_CVPR}.

\textbf{PKU-MMD}.\;
PKU-MMD \cite{liu2017pku} is a multi-modal RGB-D benchmark for human action understanding captured with Kinect v2 sensors. Following prior ZSSAR protocols \cite{9506179, zhou2023zero, Zhu_2024_CVPR}, we use the commonly adopted 51-class setting, which contains 1,076 long video sequences and nearly 20,000 action instances performed by 66 subjects from three camera views. The standard evaluation protocols are cross-subject (Xsub) and cross-view (Xview). 

\textbf{Kinetics-200/400}.\;
Kinetics-400 \cite{kay2017kinetics} is a large-scale in-the-wild video action recognition dataset collected from YouTube videos. Unlike the NTU-RGB-D 60/120 and PKU-MMD datasets above, Kinetics contains larger variations in scene, viewpoint, subject appearance, camera motion, and action context. Following the PURLS benchmark \cite{Zhu_2024_CVPR}, Kinetics-200 is constructed from the first 200 classes of Kinetics-400 and evaluated under multiple seen/unseen splits. We additionally evaluate on Kinetics-400 using analogous splits to test scalability under a larger action vocabulary. 

\textbf{Skeleton extraction}.\;
For all datasets, we extract 2D skeletons from RGB frames using RTMPose \cite{jiang2023rtmpose}. We also obtain pose-anchored semantic cues from the same pose estimation process.

\textbf{Auxiliary HPE training data}.\;
We train the 2D pose estimator on MS COCO \cite{lin2014microsoft}. In addition to the original pose estimation supervision, we use image-level textual descriptions from COCO captions as semantic supervision for the proposed action-semantic alignment loss. Specifically, the pose-anchored semantic cue extracted from the HPE representation is aligned with the corresponding caption feature encoded by the pretrained CLIP text encoder \cite{radford2021learning}. MS COCO is used only for training the HPE model and is not used for ZSSAR training.

\subsection{Seen/Unseen Split Details}
\label{app:split_details}
We follow the seen/unseen class splits used in prior ZSSAR protocols for fair comparison. 
For NTU-RGB+D 60, we evaluate the 55/5 and 48/12 splits; for NTU-RGB+D 120, we evaluate the 110/10 and 96/24 splits. 
For PKU-MMD, we use the standard 46/5 split adopted in previous ZSSAR works. 
For Kinetics-200, we follow the PURLS benchmark~\cite{Zhu_2024_CVPR} and evaluate the 180/20, 160/40, 140/60, and 120/80 splits. For Kinetics-400, we use analogous 360/40, 320/80, 300/100, and 280/120 splits. All models are trained only on the seen classes of each split, and no samples from unseen classes are used during training.

\subsection{Action Description Details}
\label{app:action_description_details}
We use LLM-generated action descriptions to construct text prototypes for ZSSAR. 
Following the semantic setting introduced by SMIE~\cite{zhou2023zero}, each action category label is expanded into a richer action description using ChatGPT. For example, the category label ``drink water'' is expanded into ``the act of taking in water by mouth to hydrate or quench thirst as a human action.'' 
These action-level descriptions are encoded by the frozen CLIP text encoder~\cite{radford2021learning} and used as semantic prototypes for both seen and unseen classes. 

Recent ZSSAR methods further exploit LLM-generated semantics by constructing additional fine-grained descriptions, such as body-part-specific, temporal-phase-specific, or local/global action descriptions~\cite{Zhu_2024_CVPR,chen2024fine,Wu_2025_ICCV}. 
In contrast, PoseBridge does not introduce extra fine-grained LLM descriptions on the text side. 
We use the same action-level descriptions following SMIE and instead recover additional action-relevant semantics from the HPE process through pose-anchored semantic cues. This design ensures that the performance gain does not come from stronger text-side description engineering, but from the proposed pose-anchored semantic bridge.
\section{Additional Performance Comparisons}
\label{app:additional_evaluation_settings}

\begin{table*}[t]
\caption{Results under more challenging seen/unseen splits on NTU-RGB+D 60 and NTU-RGB+D 120. 
The best results are highlighted in \best{red}, and the second-best results are \second{blue}.}
\label{tab:challenging_splits}
\vspace{3pt}
\centering
\setlength{\tabcolsep}{6.2pt}
\renewcommand{\arraystretch}{1.15}

\resizebox{0.8\textwidth}{!}{%
\begin{tabular}{lc!{\datasetsep}
c!{\splitsep} c!{\datasetsep}
c!{\splitsep} c}
\toprule

\multirow{2}{*}{\textbf{Method}}
& \multirow{2}{*}{\textbf{Venue}}
& \multicolumn{2}{c!{\datasetsep}}{\textbf{NTU-RGB+D 60 (Xsub)}}
& \multicolumn{2}{c}{\textbf{NTU-RGB+D 120 (Xsub)}} \\

\cmidrule(lr){3-4}
\cmidrule(lr){5-6}

&
& 40/20 Split
& 30/30 Split
& 80/40 Split
& 60/60 Split \\

\midrule

ReViSE \cite{Tsai_2017_ICCV} & ICCV 2017
& 24.3 & 14.8 & 19.5 & 8.3 \\

JPoSE \cite{Wray_2019_ICCV} & ICCV 2019
& 20.1 & 12.4 & 13.7 & 7.7 \\

CADA-VAE \cite{Schonfeld_2019_CVPR_Workshops} & CVPR 2019
& 16.2 & 11.5 & 10.6 & 5.7 \\

SynSE \cite{9506179} & ICIP 2021
& 19.9 & 12.0 & 13.6 & 7.7  \\

PURLS \cite{Zhu_2024_CVPR} & CVPR 2024
& 31.1 & 23.5 & 28.4 & 19.6 \\

SCoPLe \cite{Zhu_2025_CVPR} & CVPR 2025
& 32.0 & 18.2 & 25.3 & 15.7 \\

TDSM \cite{Do_2025_ICCV} & ICCV 2025
& \second{36.1} & \second{25.9} & \second{37.0} & \second{27.2} \\

\rowcolor{softlavender}
\textbf{PoseBridge} & \textbf{This work}
& \best{41.4} & \best{37.2} & \best{43.7} & \best{30.4} \\

\bottomrule
\end{tabular}%
}
\end{table*}

\begin{table*}[t]
\caption{Comparison on NTU-RGB+D 60 under the Xview protocol and NTU-RGB+D 120 under the Xset protocol (standard seen/unseen splits). 
We report ZSL accuracy and GZSL seen accuracy $S$, unseen accuracy $U$, and harmonic mean $H$.}
\label{tab:xview_xset}
\vspace{3pt}
\centering
\setlength{\tabcolsep}{4.2pt}
\renewcommand{\arraystretch}{1.15}

\resizebox{\textwidth}{!}{%
\begin{tabular}{lc!{\datasetsep}
cccc!{\splitsep}
cccc!{\datasetsep}
cccc!{\splitsep}
cccc}
\toprule

\multirow{4}{*}{\textbf{Method}} 
& \multirow{4}{*}{\textbf{Venue}}
& \multicolumn{8}{c!{\datasetsep}}{\textbf{NTU-RGB+D 60 (Xview)}}
& \multicolumn{8}{c}{\textbf{NTU-RGB+D 120 (Xset)}} \\

\cmidrule(lr){3-10}\cmidrule(lr){11-18}

& 
& \multicolumn{4}{c!{\splitsep}}{55/5 Split}
& \multicolumn{4}{c!{\datasetsep}}{48/12 Split}
& \multicolumn{4}{c!{\splitsep}}{110/10 Split}
& \multicolumn{4}{c}{96/24 Split} \\

\cmidrule(lr){3-6}
\cmidrule(lr){7-10}
\cmidrule(lr){11-14}
\cmidrule(lr){15-18}

&
& \multicolumn{1}{c}{ZSL}
& \multicolumn{3}{c!{\splitsep}}{GZSL}
& \multicolumn{1}{c}{ZSL}
& \multicolumn{3}{c!{\datasetsep}}{GZSL}
& \multicolumn{1}{c}{ZSL}
& \multicolumn{3}{c!{\splitsep}}{GZSL}
& \multicolumn{1}{c}{ZSL}
& \multicolumn{3}{c}{GZSL} \\

\cmidrule(lr){3-3}\cmidrule(lr){4-6}
\cmidrule(lr){7-7}\cmidrule(lr){8-10}
\cmidrule(lr){11-11}\cmidrule(lr){12-14}
\cmidrule(lr){15-15}\cmidrule(lr){16-18}

&
& \textit{Acc} & \textit{S} & \textit{U} & \textit{H}
& \textit{Acc} & \textit{S} & \textit{U} & \textit{H}
& \textit{Acc} & \textit{S} & \textit{U} & \textit{H}
& \textit{Acc} & \textit{S} & \textit{U} & \textit{H} \\

\midrule

ReViSE \cite{Tsai_2017_ICCV} & ICCV 2017
& 54.4 & 25.8 & 29.3 & 27.4
& 17.2 & 34.2 & 16.4 & 22.1
& 30.2 & 4.0 & 23.7 & 6.8
& 13.5 & 2.6 & 3.4 & 2.9 \\

JPoSE \cite{Wray_2019_ICCV} & ICCV 2019
& 72.0 & 61.1 & 59.5 & 60.3 
& 28.9 & 29.0 & 14.7 & 19.5 
& 52.8 & 23.6 & 4.4 & 7.4 
& 38.5 & 79.3 & 2.6 & 4.9 \\

CADA-VAE \cite{Schonfeld_2019_CVPR_Workshops} & CVPR 2019
& 75.1 & 65.7 & 56.1 & 60.5 
& 32.9 & 49.7 & 25.9 & 34.0 
& 52.5 & 46.0 & 44.5 & 45.2 
& 38.7 & 47.6 & 26.8 & 34.3 \\

SynSE \cite{9506179} & ICIP 2021
& 68.0 & 65.5 & 45.6 & 53.8 
& 29.9 & 61.3 & 24.6 & 35.1  
& 59.3 & 58.9 & 49.2 & 53.6  
& 41.4 & 46.8 & 31.8 & 37.9  \\

SMIE \cite{zhou2023zero} & ACMMM 2023
& 79.0 & - & - & -
& 41.0 & - & - & -
& 57.0 & - & - & -
& 42.3 & - & - & - \\

STAR \cite{chen2024fine} & ACMMM 2024
& \second{81.6} & 71.9 & 70.3 & 71.1 
& 42.5 & 66.2 & 37.5 & 47.9 
& 65.3 & 59.3 & 59.5 & 59.4 
& 44.1 & 53.7 & 34.1 & 41.7 \\

Neuron \cite{Chen_2025_CVPR} & CVPR 2025
& \best{87.8} & 70.6 & 75.9 & \second{73.2}
& \second{63.3} & 65.3 & 58.1 & \second{61.5}
& \second{71.1} & 67.5 & 58.9 & \second{62.9}
& \second{54.0} & 67.0 & 44.9 & \second{53.8} \\

\rowcolor{softlavender}
\textbf{PoseBridge} & \textbf{This work}
& \best{87.8} & 83.9 & 84.1 & \best{84.0}
& \best{65.9} & 73.7 & 57.1 & \best{64.4} 
& \best{77.1} & 70.3 & 64.5 & \best{67.3}
& \best{65.0} & 67.9 & 55.1 & \best{60.8} \\

\bottomrule
\end{tabular}%
}
\end{table*}

\subsection{Results under More Challenging Seen/Unseen Splits}
\label{app:challenging_splits}
We further evaluate PoseBridge under more challenging seen/unseen splits, where the number of unseen classes is substantially increased. As summarized in Table~\ref{tab:challenging_splits}, PoseBridge consistently achieves the best performance across all settings. The improvement is especially pronounced on the most challenging NTU-RGB+D 60 30/30 split, where PoseBridge improves the previous best result from 25.9\% to 37.2\%. These results indicate that recovering pose-anchored semantics is particularly useful when the model must distinguish among a larger and more diverse set of unseen action categories.

\subsection{Results on Xview and Xset Protocols}
\label{app:xview_xset_results}
We also test whether PoseBridge generalizes beyond the Xsub protocols used in the main paper. Table~\ref{tab:xview_xset} reports additional results on NTU-RGB+D 60 Xview and NTU-RGB+D 120 Xset. 
PoseBridge achieves strong performance across both protocols and remains competitive or superior in both ZSL and GZSL settings. These results confirm that the proposed bridge is not limited to subject-based generalization, but also remains effective under viewpoint and setup shifts.

\begin{table*}[t]
\caption{
Controlled comparison under the same estimated 2D skeleton input setting. 
All methods use the same RTMPose-extracted COCO-17 skeletons as input. 
We report ZSL accuracy and GZSL harmonic mean $H$ on NTU-RGB+D 60/120 standard splits.
}
\label{tab:estimated_2d_sota_comparison}
\vspace{3pt}
\centering
\setlength{\tabcolsep}{5.2pt}
\renewcommand{\arraystretch}{1.15}

\resizebox{\textwidth}{!}{%
\begin{tabular}{lc!{\datasetsep}
cc!{\splitsep}cc!{\datasetsep}
cc!{\splitsep}cc}
\toprule

\multirow{3}{*}{\textbf{Method}}
& \multirow{3}{*}{\textbf{Venue}}
& \multicolumn{4}{c!{\datasetsep}}{\textbf{NTU-RGB+D 60 (Xsub)}}
& \multicolumn{4}{c}{\textbf{NTU-RGB+D 120 (Xsub)}} \\

\cmidrule(lr){3-6}
\cmidrule(lr){7-10}

&
& \multicolumn{2}{c!{\splitsep}}{55/5 Split}
& \multicolumn{2}{c!{\datasetsep}}{48/12 Split}
& \multicolumn{2}{c!{\splitsep}}{110/10 Split}
& \multicolumn{2}{c}{96/24 Split} \\

\cmidrule(lr){3-4}
\cmidrule(lr){5-6}
\cmidrule(lr){7-8}
\cmidrule(lr){9-10}

&
& \;\;ZSL & GZSL \textit{(H)}
& \;\;ZSL & GZSL \textit{(H)}
& \;\;ZSL & GZSL \textit{(H)}
& \;\;ZSL & GZSL \textit{(H)} \\

\midrule

Neuron~\cite{Chen_2025_CVPR} & CVPR 2025
& \;\;\second{88.3} & \second{61.1}
& \;\;58.8 & \second{49.0}
& \;\;66.5 & \second{48.2}
& \;\;\second{60.2} & 45.6 \\

TDSM~\cite{Do_2025_ICCV} & ICCV 2025
& \;\;86.9 & -
& \;\;\second{64.7} & -
& \;\;\second{76.3} & -
& \;\;59.7 & - \\

FS-VAE~\cite{Wu_2025_ICCV} & ICCV 2025
& \;\;87.8 & 32.7
& \;\;48.8 & 41.8
& \;\;62.6 & 46.9
& \;\;56.2 & \second{48.5} \\

\rowcolor{softlavender}
\textbf{PoseBridge} & \textbf{This work}
& \;\;\best{88.8} & \best{83.0}
& \;\;\best{73.2} & \best{68.0}
& \;\;\best{80.3} & \best{68.0}
& \;\;\best{70.9} & \best{62.6} \\

\bottomrule
\end{tabular}%
}
\end{table*}

\subsection{Fair Comparison with Estimated 2D Skeletons}
\label{app:estimated_2d_skeletons}
Most prior ZSSAR methods report results using Kinect-based 3D skeletons with 25 body joints, whereas PoseBridge uses estimated 2D skeletons in the COCO-17 format extracted by RTMPose \cite{jiang2023rtmpose}. To ensure a fair comparison under the same skeleton input setting, we re-evaluate representative recent methods, including Neuron \cite{Chen_2025_CVPR}, TDSM \cite{Do_2025_ICCV}, and FS-VAE \cite{Wu_2025_ICCV}, using the same RTMPose-extracted 2D skeletons as PoseBridge. 

As shown in Table~\ref{tab:estimated_2d_sota_comparison}, PoseBridge remains consistently superior under this controlled setting. This confirms that the improvement does not come from an input-format advantage, but from the proposed pose-anchored semantic preservation and transfer mechanism. Even with the same estimated 2D skeleton inputs, PoseBridge effectively recovers action-relevant cues lost during skeleton extraction and transfers them to the semantic prototype space. This further demonstrates that our gains cannot be explained by differences in skeleton input formats alone, but benefit from the proposed pose-anchored semantic preservation and transfer.

\begin{table*}[t]
\caption{HPE-stage pose estimation quality and efficiency. We compare the standard RTMPose-s with the PoseBridge-enhanced RTMPose-s used to extract both skeletons and pose-anchored semantics. AP is evaluated on the MS COCO validation set, while parameters, GFLOPs, and FPS are measured for the HPE model.}
\label{tab:hpe_pose_quality}
\vspace{3pt}
\centering
\setlength{\tabcolsep}{6pt}
\renewcommand{\arraystretch}{1.15}

\resizebox{0.95\textwidth}{!}{%
\begin{tabular}{lccccc}
\toprule
\textbf{Model} 
& \textbf{Training objective} 
& \textbf{AP} 
& \textbf{\#Params (M)} 
& \textbf{GFLOPs (G)} 
& \textbf{FPS} \\
\midrule

RTMPose-s
& $\mathcal{L}_{\mathrm{pose}}$ 
& 70.86 
& 5.47 & 0.68 & 4740.71 \\

RTMPose-s + PoseBridge 
& $\mathcal{L}_{\mathrm{pose}} + \mathcal{L}_{\mathrm{sem}}^{\mathrm{HPE}}$ 
& 70.24 
& 8.85 & 0.96 & 2534.7 \\

\bottomrule
\end{tabular}%
}
\end{table*}

\subsection{Effect of Semantic Alignment on Pose Estimation}
\label{app:hpe_pose_quality_analysis}
Since PoseBridge modifies the HPE model by adding semantic alignment during HPE training, we examine its effect on both pose estimation quality and HPE-stage efficiency. We compare the standard RTMPose-s setting with our HPE model trained using the action-semantic alignment objective in Sec.~\ref{sec:method_HPE}. 
We evaluate pose estimation quality using mean Average Precision (AP) on the MS COCO~\cite{lin2014microsoft} validation set, and report the parameters, GFLOPs, and FPS of the pose estimator.

Different from Table~\ref{tab:efficiency_comparison}, which measures only the ZSSAR recognition stage after skeleton extraction, Table~\ref{tab:hpe_pose_quality} isolates the HPE cost. As shown in Table~\ref{tab:hpe_pose_quality}, our HPE model maintains comparable pose estimation quality to the original RTMPose-s~\cite{jiang2023rtmpose}, with AP changing only from 70.86 to 70.24. The added semantic alignment increases the HPE-stage cost moderately, from 0.68 to 0.96 GFLOPs, while preserving real-time throughput. This indicates that PoseBridge can learn pose-anchored semantic cues without substantially degrading pose estimation quality or HPE efficiency. Thus, the downstream ZSSAR gains are not caused by an improved skeleton format, but by more semantically useful HPE representations.
\section{Additional Ablation and Design Studies}
\label{app:design_analysis}

\begin{figure*}[t]
\begin{center}
\includegraphics[width=1.0\textwidth]{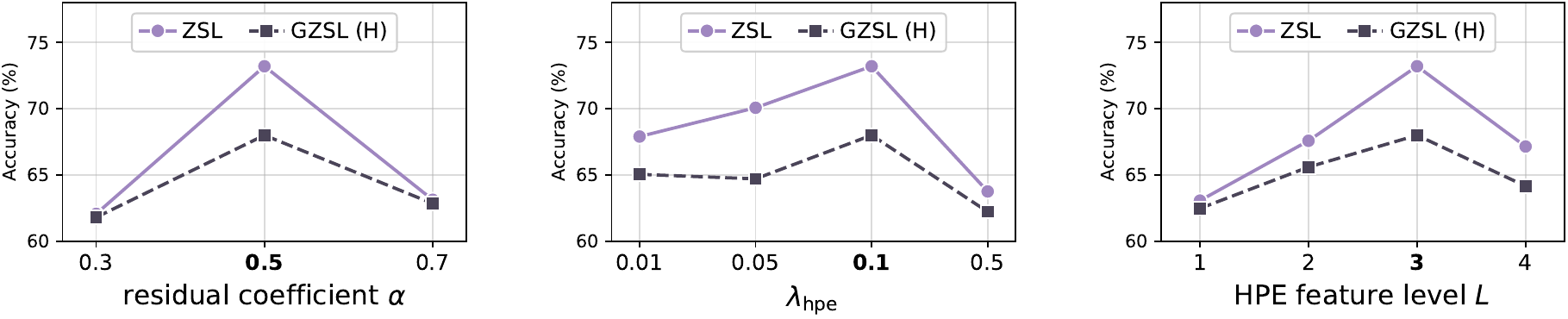}
\end{center}
\caption{HPE-side hyperparameter analysis on NTU-RGB+D 60 under the standard 48/12 split. We vary the hierarchical refinement residual coefficient $\alpha$, the semantic alignment weight $\lambda_{\mathrm{hpe}}$, and the number of HPE feature levels $L$, while keeping the other settings fixed.}
\label{fig:hpe_hyperparameters}
\end{figure*}

\begin{figure*}[t]
\begin{center}
\includegraphics[width=1.0\textwidth]{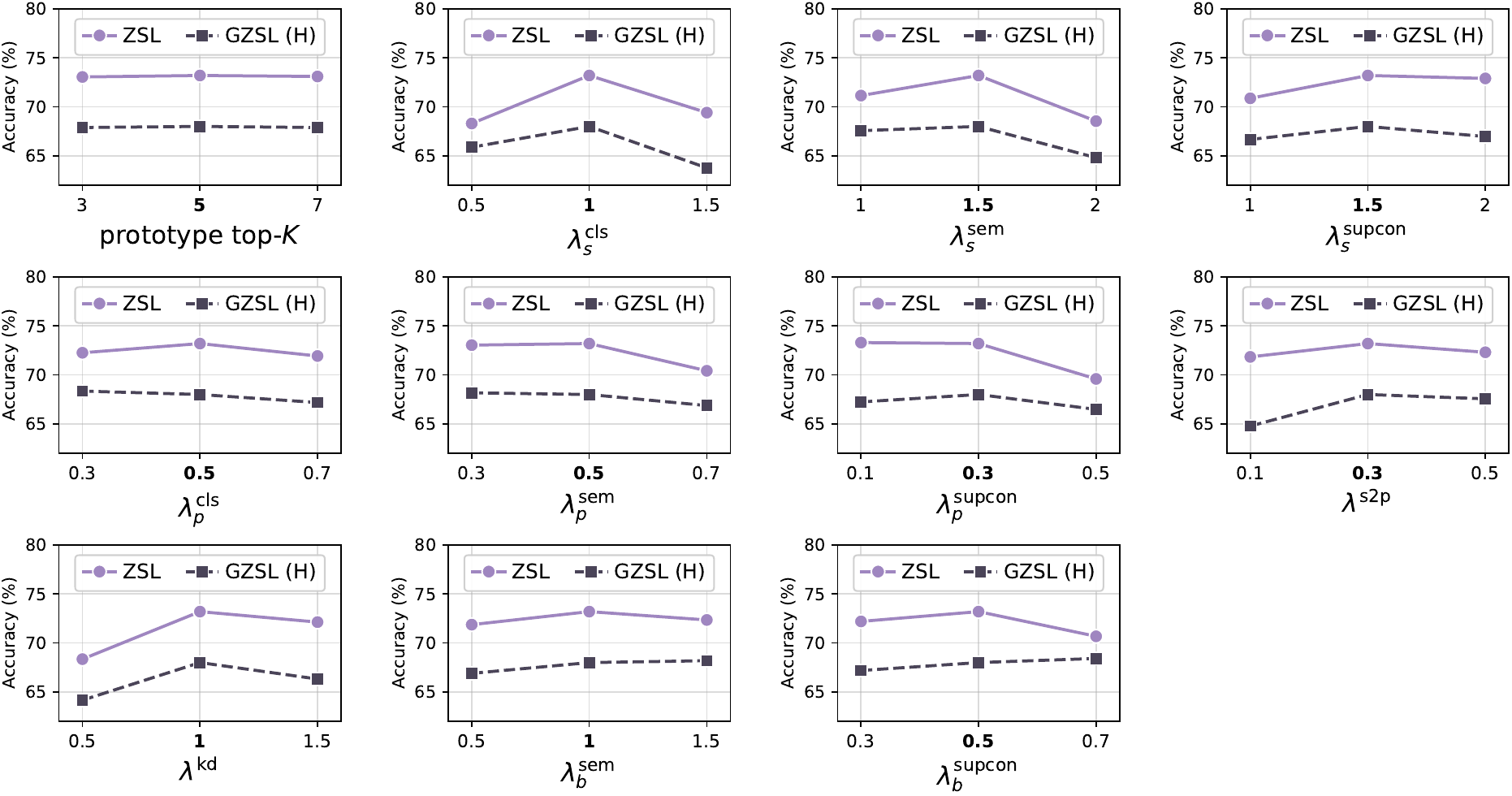}
\end{center}
\caption{ZSSAR-side hyperparameter analysis on NTU-RGB+D 60 under the standard 48/12 split. We vary prototype top-$K$ and key loss weights $\lambda$ while keeping the other settings fixed.}
\label{fig:zssar_hyperparameters}
\end{figure*}

\subsection{HPE-side Hyperparameters}
\label{app:hpe_hyperparameters}
We analyze the HPE-side hyperparameters used in Sec.~\ref{sec:method_HPE}. Specifically, we vary the residual coefficient $\alpha$ in hierarchical refinement, the semantic alignment weight $\lambda_{\mathrm{hpe}}$, and the number of HPE feature levels $L$. As shown in Fig.~\ref{fig:hpe_hyperparameters}, the default setting $\alpha=0.5$, $\lambda_{\mathrm{hpe}}=0.1$, and $L=3$ provides the best overall performance. Using a small $\lambda_{\mathrm{hpe}}$ weakens semantic alignment, while an overly large value can disturb pose-oriented representation learning. Similarly, using too few HPE feature levels limits access to fine-grained visual cues, whereas adding more levels may introduce less action-relevant low-level responses. These results support our default HPE design for extracting pose-anchored semantic cues.

\subsection{ZSSAR-side Hyperparameters}
\label{app:zssar_hyperparameters}
We further analyze the ZSSAR-side hyperparameters used in Sec.~\ref{sec:method_ZSSAR} and Sec.~\ref{sec:training_inference}. Figure~\ref{fig:zssar_hyperparameters} shows the sensitivity to prototype adaptation, skeleton-side losses, pose-semantic losses, cross-branch regularization, and bridge-side losses. Overall, PoseBridge is stable across a reasonable range of hyperparameters, and the default values used in our experiments generally provide the best or near-best performance. The model is relatively insensitive to the prototype top-$K$, while loss weights related to semantic alignment and cross-branch transfer show clearer effects. This is consistent with our motivation: the final performance depends not only on skeleton-text matching, but also on how effectively pose-anchored semantics are transferred to the skeleton representation.

\begin{table}[t]
\caption{Component-wise computational profiles of PoseBridge. The upper block reports individual module costs, while the lower block shows cumulative configurations from the skeleton-only baseline to the full model. Detector and HPE costs are aggregated over 16-frame clips, and throughput is measured on a single NVIDIA RTX A6000.}
\label{tab:component_cost}
\vspace{3pt}
\centering
\small
\setlength{\tabcolsep}{4.5pt}
\renewcommand{\arraystretch}{1.12}

\begin{tabular}{@{}
p{0.42\linewidth}
C{0.17\linewidth}
C{0.17\linewidth}
C{0.17\linewidth}
@{}}
\toprule
\textbf{Component} 
& \textbf{\#Params (M)} 
& \textbf{GFLOPs (G)} 
& \textbf{Throughput (clips/s)} \\
\midrule

\rowcolor{gray!12}
\multicolumn{4}{@{}l@{}}{\textbf{Individual module profiles}} \\
\quad RTMDet-nano
& 0.99 & 5.12 & 550.5 \\
\quad RTMPose-s
& 5.47 & 10.91 & 723.0 \\
\quad Shift-GCN
& 0.8 & 0.778 & 3536.8 \\
\quad CLIP Text Encoder
& 123.7 & 6.548 & -- \\

\midrule

\rowcolor{gray!12}
\multicolumn{4}{@{}l@{}}{\textbf{Cumulative pipeline configurations}} \\
\quad Baseline 
& 131.95 & 23.47 & 238.6 \\
\quad +HR
& 135.33 & 27.92 & 234.5 \\
\quad +HR+BP
& 135.33 & 27.92 & 233.7 \\
\quad +HR+BP+SB
& 138.05 & 27.94 & 233.0 \\

\rowcolor{softlavender}
\quad +HR+BP+SB+PA (Full)
& 138.05 & 27.94 & 232.8 \\

\bottomrule
\end{tabular}
\end{table}

\subsection{Component-wise Computational Cost}
\label{app:component_cost}
We analyze the computational cost of PoseBridge in Table~\ref{tab:component_cost}. The upper block profiles individual modules, including RTMDet-nano, RTMPose-s, Shift-GCN, and the CLIP text encoder. The lower block reports cumulative pipeline configurations, progressively adding HR, BP, SB, and PA to the skeleton-only baseline. Under the 16-frame profiling setting, the full model increases the reported complexity from 23.47 to 27.94 GFLOPs while maintaining comparable throughput. Most additional computation arises from HPE-side semantic preservation, whereas BP and PA introduce negligible overhead. These results highlight the efficiency of exploiting the existing HPE stream without a separate RGB action-recognition backbone.

\begin{table*}[t]
\caption{
Analysis with different HPE settings and 2D pose estimators. 
RTMPose (original) uses estimated skeletons without HPE intermediate features or pose-anchored semantic cues. For PoseBridge-HRNet, PoseBridge-ViTPose, and PoseBridge-RTMPose, we apply the same PoseBridge framework and report HPE performance, ZSSAR performance, parameters, GFLOPs, and FPS.
}
\label{tab:different_pose_estimators}
\vspace{3pt}
\centering
\setlength{\tabcolsep}{4.5pt}
\renewcommand{\arraystretch}{1.15}

\resizebox{\textwidth}{!}{%
\begin{tabular}{l!{\datasetsep}
c!{\datasetsep}
C{1.55cm}C{1.95cm}!{\splitsep}
C{1.55cm}C{1.95cm}!{\datasetsep}
c!{\splitsep}
c!{\splitsep}
c}
\toprule

\multirow{3}{*}{\textbf{HPE Setting}}
& \multirow{3}{*}{\shortstack{\textbf{MS COCO}\\\textbf{(AP)}}}
& \multicolumn{2}{c!{\splitsep}}{\textbf{NTU-RGB+D 60 (Xsub)}}
& \multicolumn{2}{c!{\datasetsep}}{\textbf{NTU-RGB+D 120 (Xsub)}}
& \multirow{3}{*}{\shortstack{\textbf{\#Params}\\\textbf{(M)}}}
& \multirow{3}{*}{\shortstack{\textbf{GFLOPs}\\\textbf{(G)}}}
& \multirow{3}{*}{\textbf{FPS}} \\

\cmidrule(lr){3-4}
\cmidrule(lr){5-6}

&
& \multicolumn{2}{c!{\splitsep}}{48/12 Split}
& \multicolumn{2}{c!{\datasetsep}}{96/24 Split}
&
&
& \\

\cmidrule(lr){3-4}
\cmidrule(lr){5-6}

&
& ZSL & GZSL \textit{(H)}
& ZSL & GZSL \textit{(H)}
&
&
& \\

\midrule

RTMPose (original)
& 70.9
& 54.7 & 52.0
& 55.1 & 52.4
& 5.5 & 0.7 & 4740.7 \\

PoseBridge-HRNet~\cite{sun2019deep}
& 70.8
& 65.3 & 63.8
& \textbf{71.0} & 61.7
& 28.8 & 7.8 & 766.9 \\

PoseBridge-ViTPose~\cite{xu2022vitpose}
& \textbf{73.7}
& 69.2 & 61.8
& 69.8 & 61.7
& 27.4 & 6.2 & 1020.6 \\

\rowcolor{softlavender}
PoseBridge-RTMPose~\cite{jiang2023rtmpose}
& 70.2
& \textbf{73.2} & \textbf{68.0}
& 70.9 & \textbf{62.6}
& \textbf{8.9} & \textbf{1.0} & \textbf{2534.7} \\

\bottomrule
\end{tabular}%
}
\end{table*}

\subsection{Different Pose Estimators}
\label{app:different_pose_estimators}
We evaluate whether PoseBridge is robust to the choice of the 2D pose estimator. We first include RTMPose-s (original) as a skeleton-only baseline, which uses estimated skeleton sequences without HPE intermediate features or pose-anchored semantic cues. In addition to our default RTMPose-s~\cite{jiang2023rtmpose}, we experiment with HRNet-w32~\cite{sun2019deep} and ViTPose-s~\cite{xu2022vitpose}. Each pose estimator is trained on MS COCO using its standard pose-estimation recipe and hyperparameters, while applying the same HPE-side action-semantic alignment objective described in Sec.~\ref{sec:method_HPE}. 
As shown in Table~\ref{tab:different_pose_estimators}, PoseBridge consistently improves over the skeleton-only baseline and remains effective across different pose estimators. We also report MS COCO AP, parameter count, GFLOPs, and FPS for reference. These results indicate that the proposed pose-anchored semantic bridge is not tied to a specific HPE backbone. 
\section{Additional Analyses}
\label{app:additional_analyses}

\begin{figure*}[t]
\begin{center}
\includegraphics[width=1.0\textwidth]{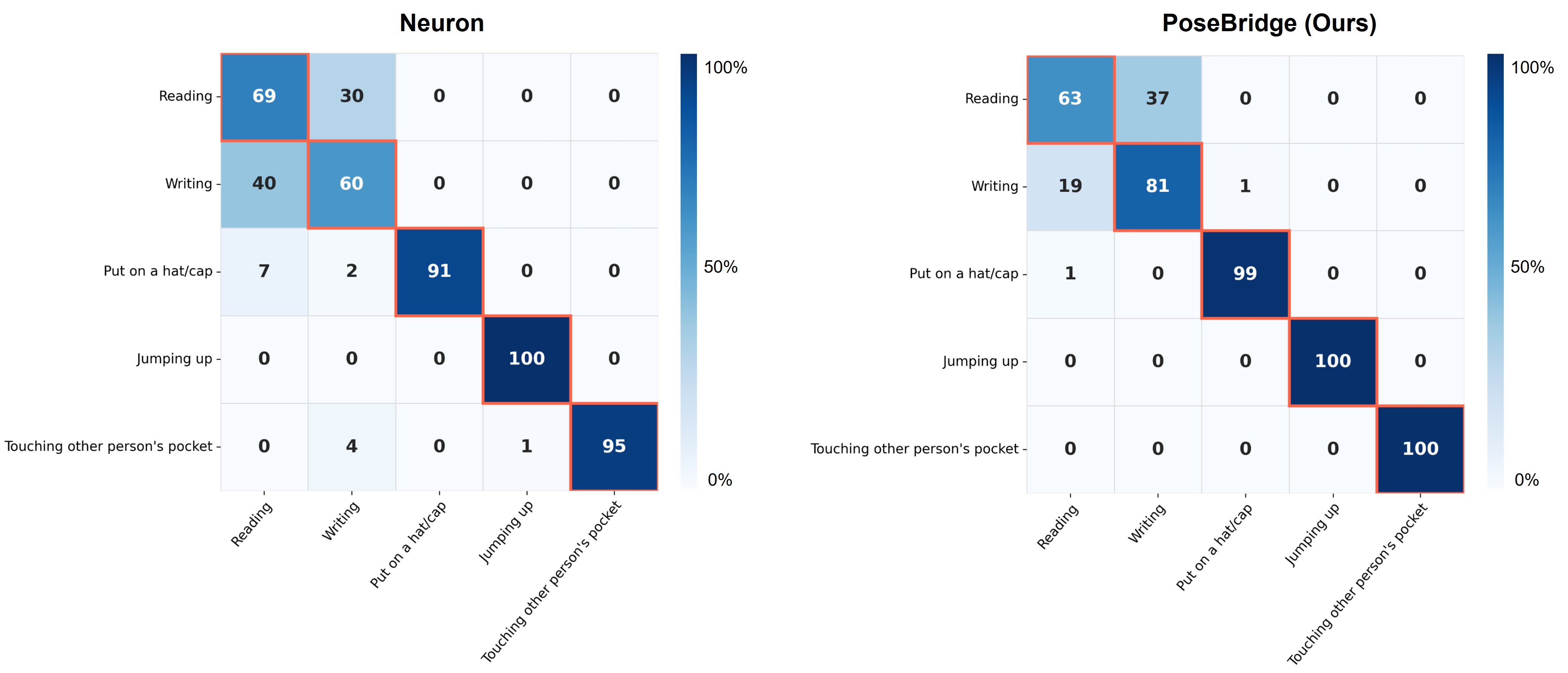}
\end{center}
\caption{Row-normalized confusion matrices of Neuron and PoseBridge on NTU-RGB+D 60 under the standard 55/5 split.}
\label{fig:cm_ntu60_55_5}
\end{figure*}

\begin{figure*}[t]
\begin{center}
\includegraphics[width=1.0\textwidth]{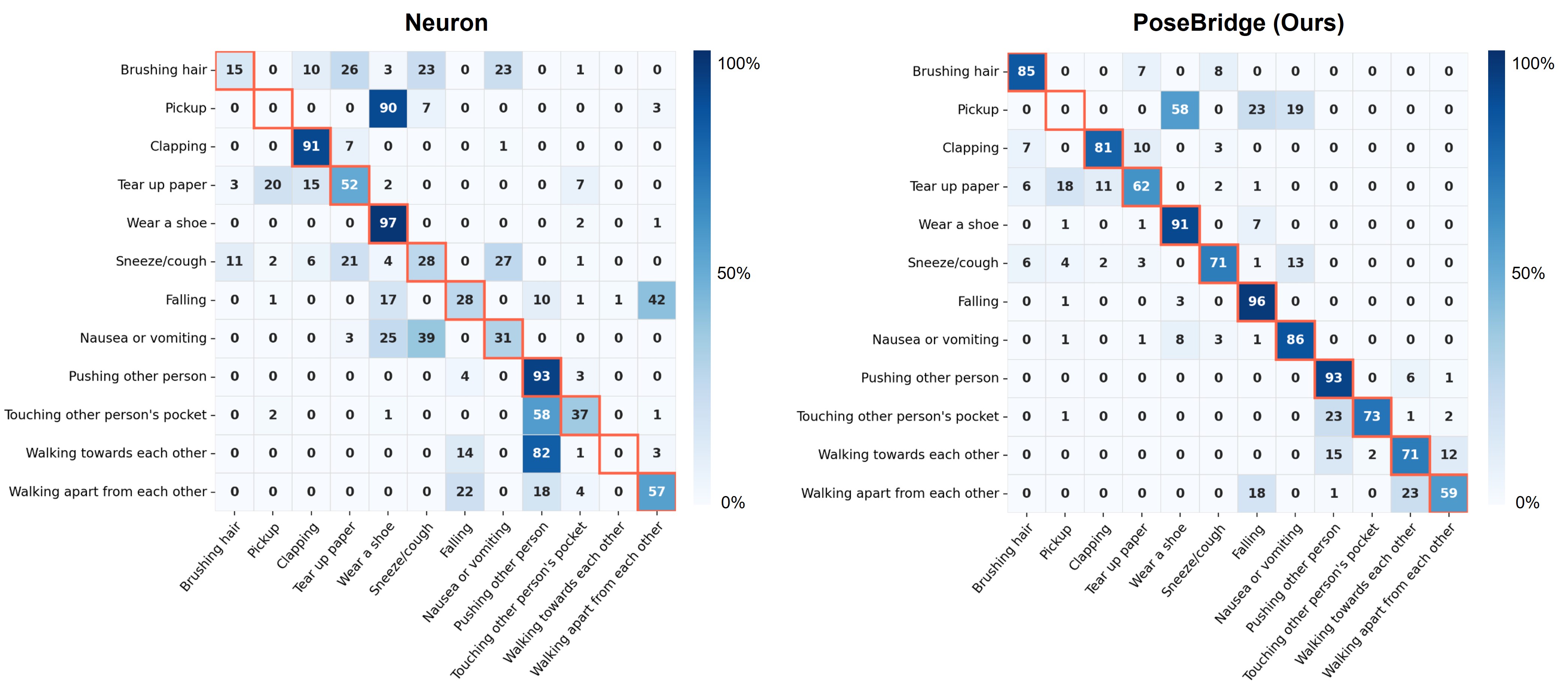}
\end{center}
\caption{Row-normalized confusion matrices of Neuron and PoseBridge on NTU-RGB+D 60 under the standard 48/12 split.}
\label{fig:cm_ntu60_48_12}
\end{figure*}

\begin{figure*}[t]
\begin{center}
\includegraphics[width=1.0\textwidth]{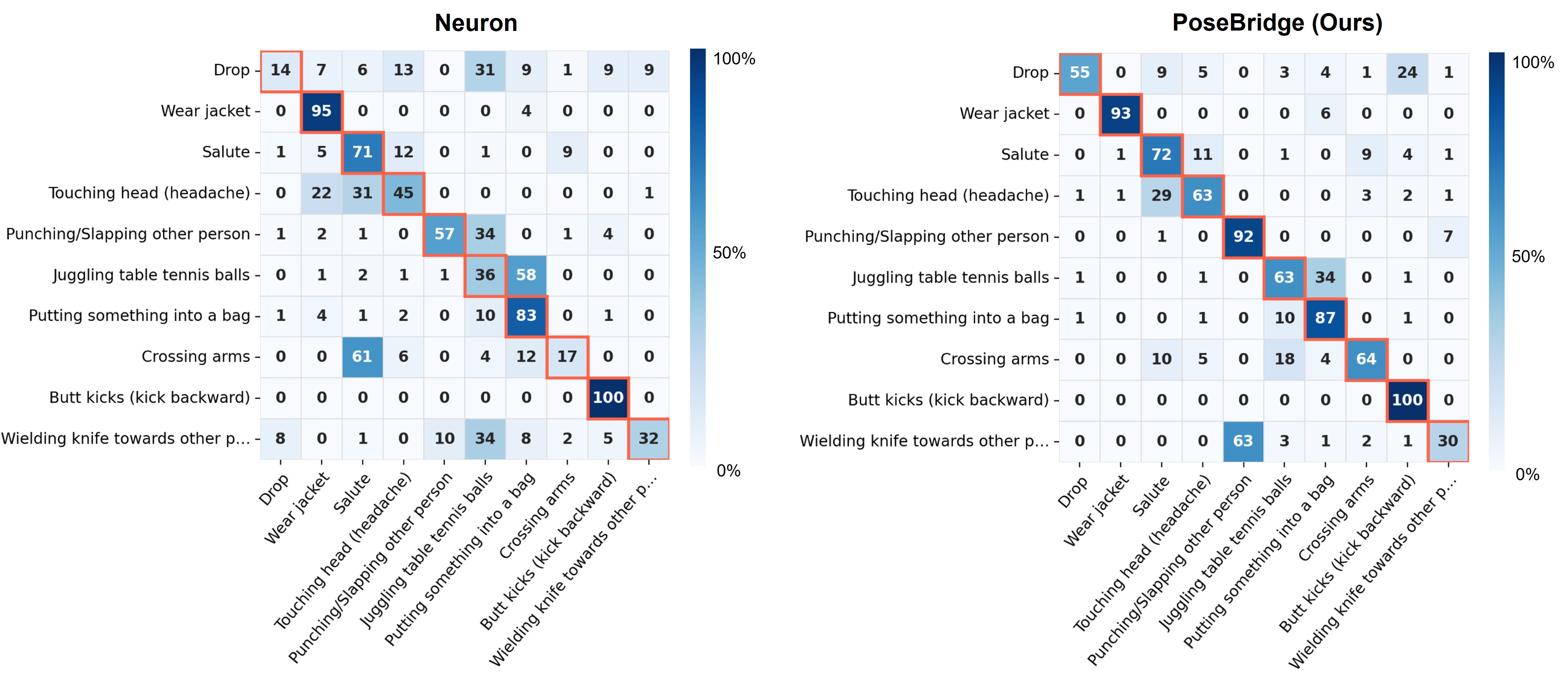}
\end{center}
\caption{Row-normalized confusion matrices of Neuron and PoseBridge on NTU-RGB+D 120 under the standard 110/10 split.}
\label{fig:cm_ntu120_110_10}
\end{figure*}

\begin{figure*}[t]
\begin{center}
\includegraphics[width=1.0\textwidth]{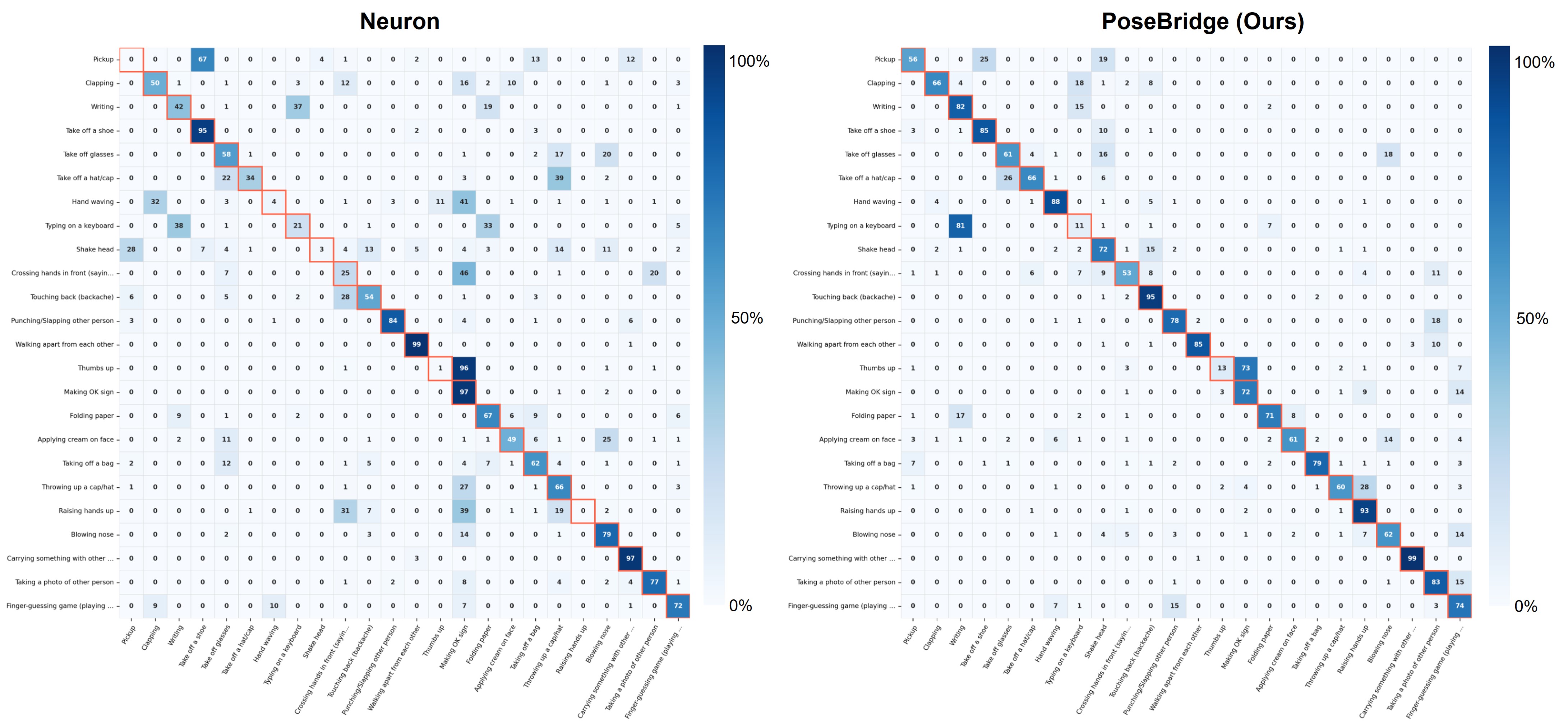}
\end{center}
\caption{Row-normalized confusion matrices of Neuron and PoseBridge on NTU-RGB+D 120 under the standard 96/24 split.}
\label{fig:cm_ntu120_96_24}
\end{figure*}

\subsection{Confusion Matrix Analysis}
\label{sec:confusion_analysis}
We provide additional confusion matrix visualizations to analyze class-level prediction behavior under the same skeleton input setting. Specifically, we compare Neuron \cite{Chen_2025_CVPR} and PoseBridge using RTMPose-extracted skeletons on the standard NTU-RGB+D 60/120 splits. As shown in Fig.~\ref{fig:cm_ntu60_55_5}--\ref{fig:cm_ntu120_96_24}, PoseBridge generally produces stronger diagonal responses and fewer off-diagonal confusions than Neuron. This trend is meaningful because both methods use the same estimated 2D skeleton inputs, indicating that the improvement comes from the proposed pose-anchored semantic bridge rather than differences in skeleton format. The results further show that pose-anchored semantics help distinguish semantically similar or motion-ambiguous actions in zero-shot recognition.

\begin{table*}[t]
\caption{Comparison with object-recognition features on NTU-RGB+D 60 (48/12). Object features are obtained from a pretrained YOLOv11 detector on person-centered crops. We report ZSL accuracy and GZSL harmonic mean (H) in percentages.}
\label{tab:object_control}
\vspace{3pt}
\centering
\setlength{\tabcolsep}{6pt}
\renewcommand{\arraystretch}{1.15}

\begin{tabular}{lcc}
\toprule
\textbf{Method}
& \textbf{ZSL}
& \textbf{GZSL (H)} \\
\midrule

Skeleton-only
& 54.7 & 52.0 \\
Skeleton + object-recognition features
& 56.6 & 56.1 \\
\textbf{PoseBridge}
& \textbf{73.2} & \textbf{68.0} \\

\bottomrule
\end{tabular}%
\end{table*}

\subsection{Object-Recognition Feature Baseline}
\label{sec:controlled_baselines}

To examine whether object recognition alone explains PoseBridge's improvements, we compare against a baseline using pretrained YOLOv11 \cite{khanam2024yolov11} object features extracted from person-centered crops and associated with pose anchors. As shown in Table~\ref{tab:object_control}, object-recognition features improve ZSL accuracy from 54.7\% to 56.6\% and GZSL-H from 52.0\% to 56.1\%, whereas PoseBridge achieves 73.2\% and 68.0\%, respectively. These results demonstrate that PoseBridge captures richer pose-anchored visual evidence beyond object identity alone, including body configuration and pose-relative interaction cues.

\begin{table*}[t]
\caption{ZSL accuracy (\%) by unseen action group on NTU-RGB+D 60 (48/12). Object/interaction-sensitive actions require additional object or interaction cues for discrimination, whereas motion-centric actions primarily depend on body motion.}
\label{tab:action_groups}
\vspace{3pt}
\centering
\setlength{\tabcolsep}{6pt}
\renewcommand{\arraystretch}{1.15}

\begin{tabular}{lcccc}
\toprule
\textbf{Unseen action group}
& \textbf{\#Cls.}
& \textbf{Skeleton-only}
& \textbf{PoseBridge}
& \textbf{Gain} \\
\midrule

Motion-centric
& 4 & 68.2 & 83.6 & +15.4 \\
Object/interaction-sensitive
& 8 & 47.9 & 67.9 & \textbf{+20.0} \\
\midrule
All unseen classes
& 12 & 54.7 & 73.2 & +18.5 \\

\bottomrule
\end{tabular}%
\end{table*}

\begin{table*}[!t]
\caption{Spatial concentration of HPE Grad-CAM activations on NTU-RGB+D 60 (48/12). Top-25\% activation mass measures the fraction of total normalized activation within the most highly activated 25\% of spatial locations.}
\label{tab:gradcam_quant}
\vspace{3pt}
\centering
\setlength{\tabcolsep}{6pt}
\renewcommand{\arraystretch}{1.15}

\begin{tabular}{lcc}
\toprule
\textbf{Metric}
& \textbf{Vanilla HPE}
& \textbf{PoseBridge-HPE} \\
\midrule
Top-25\% activation mass
& 0.387 & \textbf{0.411} \\

\bottomrule
\end{tabular}%
\end{table*}

\begin{table*}[!t]
\caption{Quantitative separability of zero-shot matching embeddings on NTU-RGB+D 60 (48/12). Metrics are computed on the original L2-normalized embeddings before t-SNE projection. Higher values indicate stronger class separation.}
\label{tab:embedding_quant}
\vspace{3pt}
\centering
\setlength{\tabcolsep}{6pt}
\renewcommand{\arraystretch}{1.15}

\begin{tabular}{lcc}
\toprule
\textbf{Embedding metric}
& \textbf{Skeleton-only}
& \textbf{PoseBridge} \\
\midrule
Silhouette score $\uparrow$
& 0.268 & \textbf{0.336} \\
Inter-/intra-class distance ratio $\uparrow$
& 2.560 & \textbf{3.295} \\
\bottomrule
\end{tabular}%
\end{table*}

\subsection{Analysis by Action Semantics}
\label{sec:action_semantics_analysis}

To investigate the benefits of pose-anchored semantics across different action types, we divide the 12 unseen classes of NTU-RGB+D 60 (48/12) into motion-centric and object/interaction-sensitive groups based on their action descriptions, before examining predictions. As shown in Table~\ref{tab:action_groups}, PoseBridge achieves larger improvements on object/interaction-sensitive actions (+20.0 points) than on motion-centric actions (+15.4 points). This supports our skeletonization-gap hypothesis that preserving visual evidence is particularly beneficial when body motion alone cannot fully distinguish semantically different actions.

\subsection{Quantitative Representation Analysis}
\label{sec:quantitative_representation}

We complement the Grad-CAM and t-SNE visualizations in Fig.~\ref{fig:qualitative_analysis} with quantitative analyses of spatial activation and embedding separability. As shown in Table~\ref{tab:gradcam_quant}, the top-25\% activation mass increases from 0.387 for vanilla HPE to 0.411 for PoseBridge-HPE, indicating more concentrated spatial responses. Moreover, Table~\ref{tab:embedding_quant} shows improvements in silhouette score (0.268 to 0.336) and inter-/intra-class distance ratio (2.560 to 3.295), measured on the original L2-normalized embeddings. These results provide quantitative support for the improved representation quality achieved by PoseBridge.

\begin{table*}[t]
\caption{Robustness evaluation on NTU-RGB+D 60 (48/12) under skeleton/HPE-feature perturbation and random RGB block occlusion. Results are ZSL accuracy (\%), and $\Delta$ denotes the change from the original input condition.}
\label{tab:robustness}
\vspace{3pt}
\centering
\setlength{\tabcolsep}{6pt}
\renewcommand{\arraystretch}{1.15}

\begin{tabular}{lcccc}
\toprule
\textbf{Input condition}
& \textbf{Skeleton}
& $\Delta$
& \textbf{PoseBridge}
& $\Delta$ \\
\midrule
Original input
& 54.7 & -- & 73.2 & -- \\
Skeleton/HPE perturbation
& 48.1 & -6.6 & 67.9 & -5.3 \\
Random RGB block occlusion
& 46.8 & -7.9 & 65.5 & -7.7 \\
\bottomrule
\end{tabular}%
\end{table*}

\begin{table*}[t]
\caption{Effect of HPE-side semantic supervision on NTU-RGB+D 60 (48/12). All variants use the same PoseBridge-ZSSAR stage. The visual distillation variant does not use paired captions during HPE adaptation, although its CLIP teacher was pretrained with image--text supervision. Results are ZSL accuracy and GZSL harmonic mean (H) in percentages.}
\label{tab:semantic_supervision}
\vspace{3pt}
\centering
\setlength{\tabcolsep}{6pt}
\renewcommand{\arraystretch}{1.15}

\begin{tabular}{lcc}
\toprule
\textbf{HPE-side supervision}
& \textbf{ZSL}
& \textbf{GZSL (H)} \\
\midrule
None (vanilla HPE features)
& 61.4 & 61.7 \\
Frozen CLIP visual teacher
& 67.0 & 64.4 \\
\textbf{COCO caption alignment (Ours)}
& \textbf{73.2} & \textbf{68.0} \\
\bottomrule
\end{tabular}%
\end{table*}

\begin{table*}[!t]
\caption{Three-seed evaluation on the standard NTU-RGB+D 60/120 Xsub splits. Results are mean $\pm$ standard deviation (\%) over three independent training seeds with fixed seen/unseen class partitions.}
\label{tab:seed_statistics}
\vspace{3pt}
\centering
\setlength{\tabcolsep}{6pt}
\renewcommand{\arraystretch}{1.15}

\begin{tabular}{lcccc}
\toprule
\textbf{Dataset (Split)}
& \textbf{ZSL}
& \textbf{GZSL (S)}
& \textbf{GZSL (U)}
& \textbf{GZSL (H)} \\
\midrule

NTU60 (55/5)
& $88.6\pm1.0$
& $81.9\pm1.0$
& $83.6\pm1.3$
& $82.7\pm1.0$ \\

NTU60 (48/12)
& $73.1\pm1.5$
& $75.2\pm1.3$
& $61.8\pm1.7$
& $67.8\pm0.9$ \\

NTU120 (110/10)
& $80.8\pm0.9$
& $71.0\pm0.8$
& $66.3\pm1.0$
& $68.6\pm0.6$ \\

NTU120 (96/24)
& $70.5\pm1.2$
& $67.9\pm1.7$
& $58.9\pm2.2$
& $62.7\pm0.9$ \\

\bottomrule
\end{tabular}%
\end{table*}

\subsection{Robustness Under Input Degradation}
\label{sec:robustness_analysis}

We evaluate PoseBridge under two controlled degradation settings on NTU-RGB+D 60 (48/12): Gaussian perturbations applied to both skeletons and HPE cues, and random block occlusions applied to RGB frames before rerunning the full pipeline. As shown in Table~\ref{tab:robustness}, PoseBridge consistently outperforms skeleton-only recognition under perturbation (67.9\% vs. 48.1\%) and occlusion (65.5\% vs. 46.8\%). These results demonstrate partial robustness to degraded observations, although severe pose-estimation failures and persistent occlusion remain challenging.

\subsection{Alternative HPE-Side Semantic Supervision}
\label{sec:alternative_supervision}

We investigate the effect of alternative semantic supervision by replacing COCO caption alignment with visual distillation from a frozen pretrained CLIP visual encoder, without using paired captions during HPE training. As shown in Table~\ref{tab:semantic_supervision}, visual distillation improves ZSL accuracy from 61.4\% to 67.0\%, while our COCO caption alignment achieves 73.2\%. Although the CLIP teacher itself is pretrained with image--text supervision, these results demonstrate the flexibility of our framework and highlight the effectiveness of explicit visual--language alignment for zero-shot transfer.

\subsection{Statistical Reliability}
\label{sec:statistical_reliability}

To assess statistical reliability, we repeat the four standard NTU-RGB+D 60/120 Xsub settings with three independent random seeds while keeping the seen/unseen class splits fixed. As shown in Table~\ref{tab:seed_statistics}, PoseBridge achieves consistent performance across runs, including 73.1$\pm$1.5\% ZSL accuracy and 67.8$\pm$0.9\% GZSL-H on NTU60 (48/12). These results confirm that the reported improvements remain stable under training randomness.

\section{Additional Qualitative Results}
\label{app:additional_qualitative_results}

\begin{figure*}[t]
\begin{center}
\includegraphics[width=1.0\textwidth]{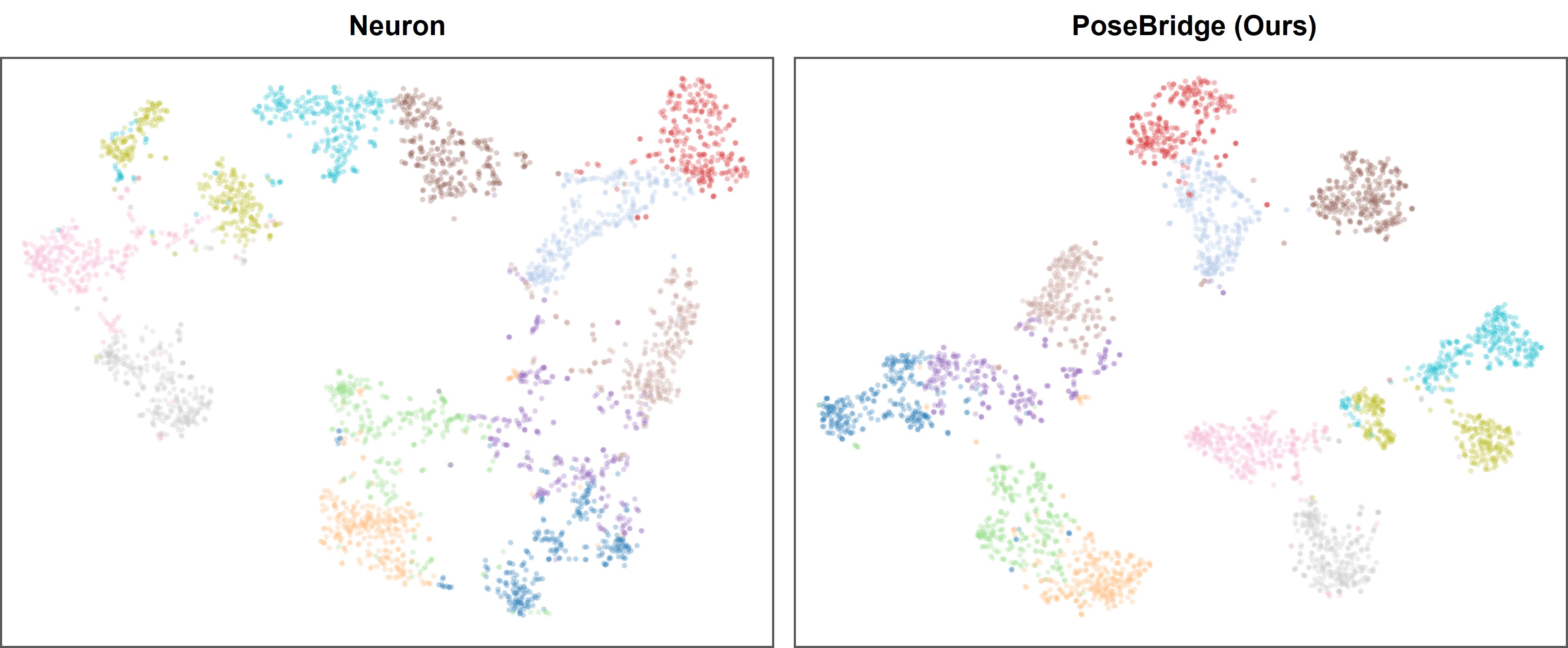}
\end{center}
\caption{t-SNE visualization of zero-shot matching features on NTU-RGB+D 60 under the standard 48/12 split. We compare Neuron and PoseBridge using the same RTMPose-extracted 2D skeleton inputs.}
\label{fig:tsne_ntu60}
\end{figure*}

\begin{figure*}[t]
\begin{center}
\includegraphics[width=1.0\textwidth]{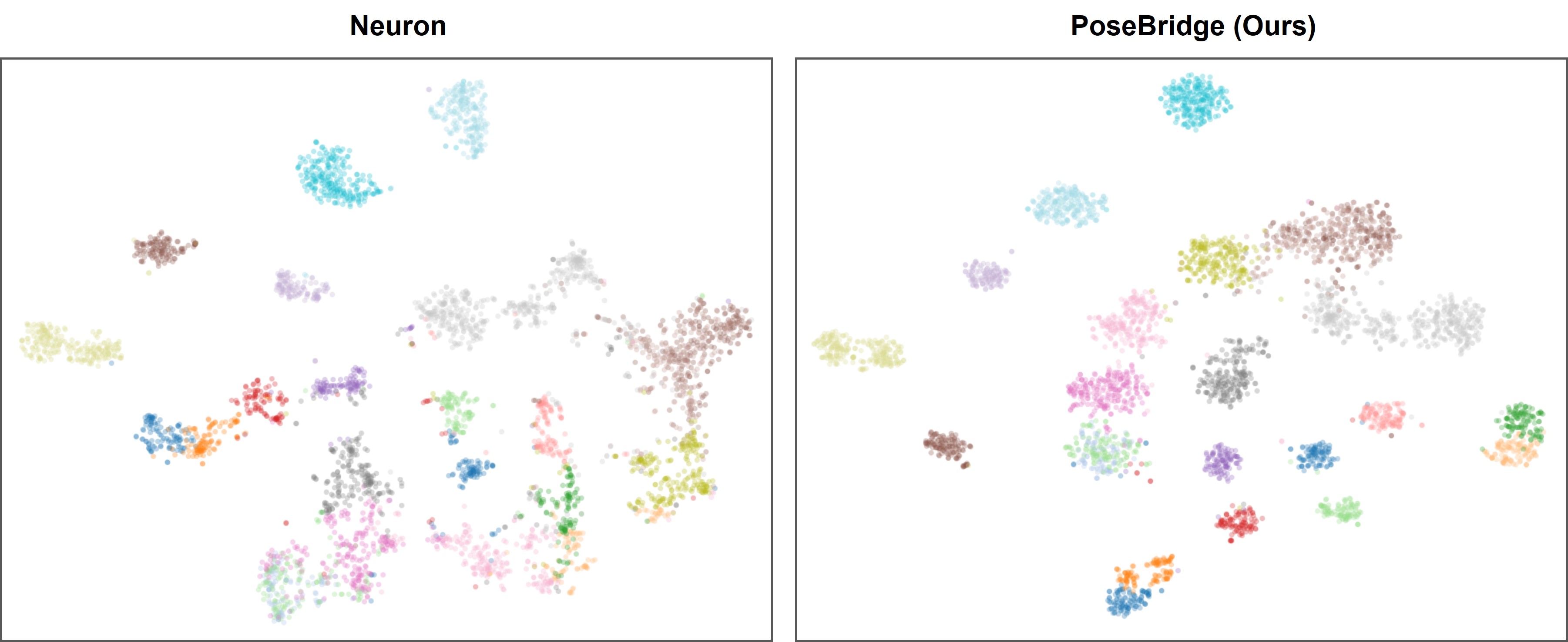}
\end{center}
\caption{t-SNE visualization of zero-shot matching features on NTU-RGB+D 120 under the standard 96/24 split. We compare Neuron and PoseBridge using the same RTMPose-extracted 2D skeleton inputs.}
\label{fig:tsne_ntu120}
\end{figure*}

\begin{figure*}[t]
\begin{center}
\includegraphics[width=1.0\textwidth]{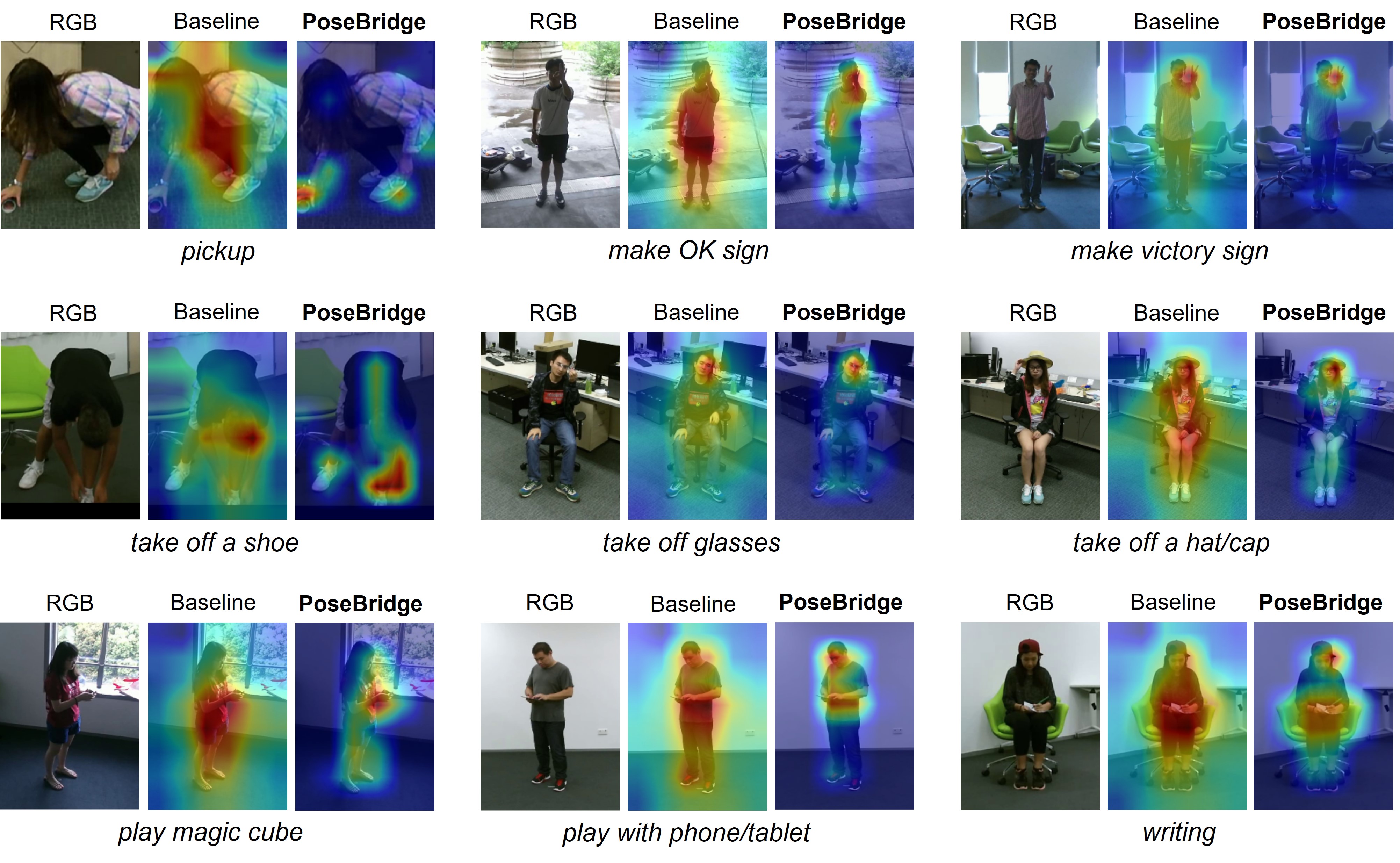}
\end{center}
\caption{Additional qualitative comparison between the baseline and PoseBridge. We show RGB samples, Grad-CAM maps of baseline HPE features, and Grad-CAM maps of PoseBridge pose-anchored semantics for various actions.}
\label{fig:additional_gradcam}
\end{figure*}

\subsection{t-SNE Visualization Against Neuron}
To further analyze the representation quality of PoseBridge, we visualize zero-shot matching features using t-SNE and compare them with Neuron~\cite{Chen_2025_CVPR}. Both methods are evaluated under the same RTMPose-extracted 2D skeleton input setting, so the comparison focuses on the learned representation space rather than differences in skeleton format. As shown in Fig.~\ref{fig:tsne_ntu60} and Fig.~\ref{fig:tsne_ntu120}, PoseBridge produces more compact intra-class clusters and clearer inter-class separation, whereas Neuron shows more scattered embeddings with noticeable overlap across classes. This indicates that the proposed pose-anchored semantics help organize skeleton features into a more discriminative semantic space. The trend is consistent on both NTU-RGB+D 60 under the 48/12 split and NTU-RGB+D 120 under the 96/24 split, suggesting that PoseBridge remains effective even when the unseen action space becomes larger and more ambiguous. These visualizations further support that bridging HPE-derived semantics to skeleton-text matching improves zero-shot class separability.

\subsection{Additional Grad-CAM Visualizations}
We provide additional Grad-CAM examples in Fig.~\ref{fig:additional_gradcam}. 
The baseline HPE model, trained only for pose estimation, tends to activate broadly over the human body, reflecting its focus on keypoint localization rather than action semantics. 
In contrast, PoseBridge produces more localized and action-discriminative responses. 
For actions such as ``pickup,'' ``take off a shoe,'' ``take off glasses,'' ``take off a hat/cap,'' ``play magic cube,'' ``play with phone/tablet,'' and ``writing,'' PoseBridge focuses more strongly on action-related visual cues around the manipulated object and the interacting body parts. 
For fine-grained gestures such as ``make OK sign'' and ``make victory sign,'' it also highlights subtle hand and finger regions that are critical for distinguishing the action. 
These results further show that PoseBridge preserves action-relevant semantics during HPE, rather than relying only on coordinate-level skeleton information, which benefits downstream zero-shot skeleton-text matching.

\section{Limitations and Broader Impacts}
\label{app:limitations_broader_impacts}

\paragraph{Limitations.}
PoseBridge reduces the skeletonization gap by reusing pose-anchored semantic cues from the same HPE process that produces skeletons. 
Its effectiveness therefore depends on the quality and semantic coverage of the HPE stream. 
When the actor is heavily occluded, the manipulated object is very small or outside the body-attended region, or the pose estimator fails under severe motion blur or crowded scenes, both the skeletons and the pose-anchored cues may become unreliable. 
In addition, PoseBridge is designed for an HPE-aware ZSSAR setting rather than a strictly coordinate-only skeleton protocol, since it assumes access to intermediate HPE representations during skeleton extraction. 
Future work could further improve robustness under challenging visual conditions and investigate lighter ways to preserve pose-relative semantic cues.

\paragraph{Broader Impacts.}
PoseBridge aims to improve zero-shot skeleton action recognition, which can reduce the need for exhaustive action-specific annotation and may benefit applications such as human-computer interaction, rehabilitation analysis, sports motion understanding, and assistive robotics. 
However, action recognition systems can also raise privacy and fairness concerns when applied to real-world videos, especially in surveillance-like or high-stakes settings. 
Although PoseBridge does not introduce an independent RGB action-recognition branch, it still relies on videos during skeleton extraction and may inherit biases from the pose estimator, caption supervision, and pretrained language models. 
We therefore recommend careful dataset-specific validation, privacy-preserving deployment, and human oversight before using such systems in sensitive applications.


\end{document}